\documentclass[sigconf]{acmart}
\usepackage{xcolor}
\usepackage{colortbl}
\usepackage{tabularx}
\usepackage{subfigure}
\usepackage[ruled]{algorithm2e} 

\bfseries
\AtBeginDocument{%
  }

\copyrightyear{2025}
\acmYear{2025}
\setcopyright{acmlicensed}\acmConference[SIGGRAPH Conference Papers '25]{Special Interest Group on Computer Graphics and Interactive Techniques Conference Conference Papers }{August 10--14, 2025}{Vancouver, BC, Canada}
\acmBooktitle{Special Interest Group on Computer Graphics and Interactive Techniques Conference Conference Papers (SIGGRAPH Conference Papers '25), August 10--14, 2025, Vancouver, BC, Canada}
\acmDOI{10.1145/3721238.3730640}
\acmISBN{979-8-4007-1540-2/2025/08}

\begin{document}

\title{SkillMimic-V2: Learning Robust and Generalizable Interaction \\ Skills from Sparse and Noisy Demonstrations}

\author{Runyi Yu}
\authornote{Both authors contributed equally to this research.}
\orcid{0009-0008-5527-7193}
\email{ingrid.yu@connect.ust.hk}
\affiliation{%
  \institution{HKUST}
  \city{Hong Kong}
  \country{China}}

\author{Yinhuai Wang}
\orcid{0000-0003-4601-4881}
\authornotemark[1]
\email{yinhuai.wang@connect.ust.hk}
\authornote{Project lead.}
\affiliation{%
  \institution{HKUST}
  \city{Hong Kong}
  \country{China}}

\author{Qihan Zhao}
\orcid{0009-0000-0377-2784}
\authornotemark[1]
\affiliation{%
 \institution{HKUST}
 \city{Hong Kong}
 \country{China}}
\email{qihan.zhao@outlook.com}

\author{Hok Wai Tsui}
\email{hwtsui@connect.ust.hk}
\orcid{0009-0002-9152-7694}
\affiliation{%
  \institution{HKUST}
  \city{Hong Kong}
  \country{China}}

\author{Jingbo Wang}
\orcid{0009-0005-0740-8548}
\email{wangjingbo1219@gmail.com}
\affiliation{%
  \institution{Shanghai AI Laboratory}
  \city{Shanghai}
  \country{China}}

\author{Ping Tan}
\authornote{Joint corresponding authors.}
\orcid{0000-0002-4506-6973}
\email{pingtan@ust.hk}
\affiliation{%
  \institution{HKUST}
  \city{Hong Kong}
  \country{China}}

\author{Qifeng Chen}
\authornotemark[3]
\orcid{0000-0003-2199-3948}
\email{cqf@ust.hk}
\affiliation{%
  \institution{HKUST}
  \city{Hong Kong}
  \country{China}}

\renewcommand{\shortauthors}{Runyi Yu, Yinhuai Wang, Qihan Zhao, Hok Wai Tusi, Jingbo Wang, Ping Tan, and Qifeng Chen}
\renewcommand{\algorithmcfname}{ALGORITHM}

\begin{abstract}
  We address a fundamental challenge in Reinforcement Learning from Interaction
Demonstration (RLID):
demonstration noise and coverage limitations. While existing data collection approaches provide valuable interaction demonstrations, they often yield sparse, disconnected, and noisy trajectories that fail to capture the full spectrum of possible skill variations and transitions.
Our key insight is that despite noisy and sparse demonstrations, there exist infinite physically feasible trajectories that naturally bridge between demonstrated skills or emerge from their neighboring states, forming a continuous space of possible skill variations and transitions. Building upon this insight, we present two data augmentation techniques: a Stitched Trajectory Graph (STG) that discovers potential transitions between demonstration skills, and a State Transition Field (STF) that establishes unique connections for arbitrary states within the demonstration neighborhood. To enable effective RLID with augmented data, we develop an Adaptive Trajectory Sampling (ATS) strategy for dynamic curriculum generation and a historical encoding mechanism for memory-dependent skill learning.
Our approach enables robust skill acquisition that significantly generalizes beyond the reference demonstrations. Extensive experiments across diverse interaction tasks demonstrate substantial improvements over state-of-the-art methods in terms of convergence stability, generalization capability, and recovery robustness.
\end{abstract}


\begin{teaserfigure}
\centering
\includegraphics[width=\linewidth]{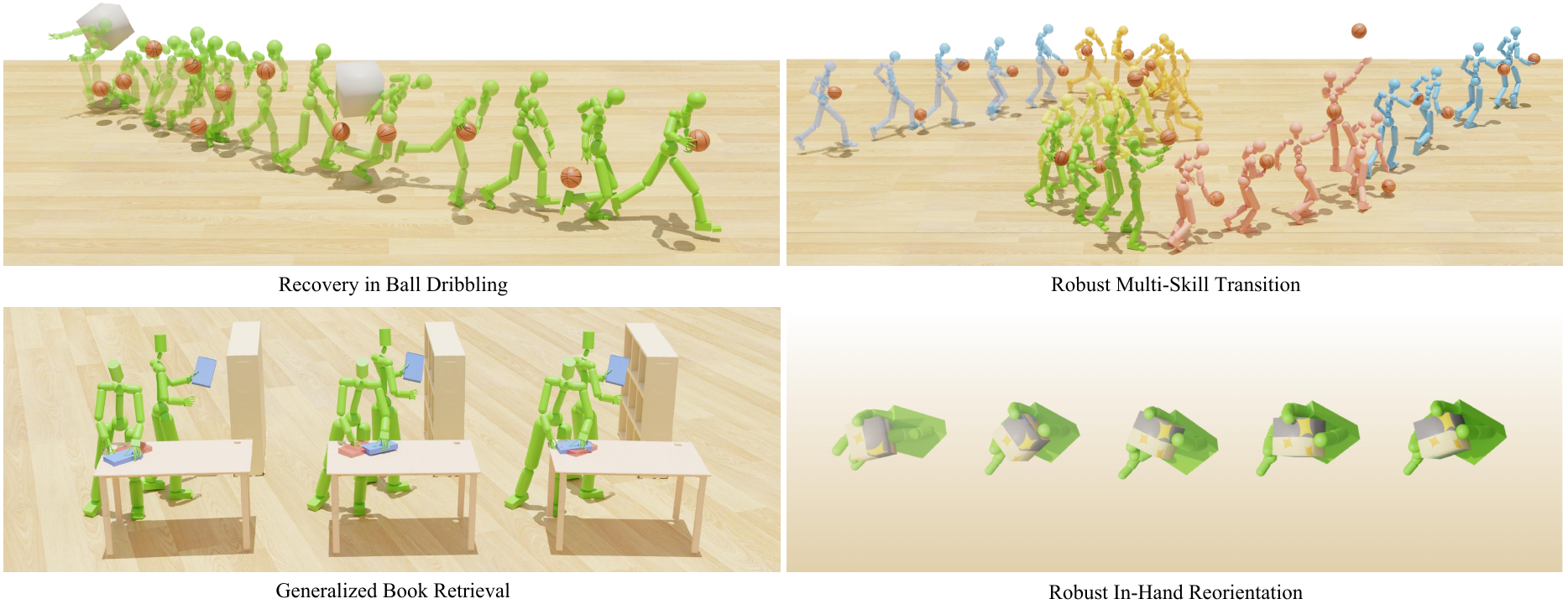}\\
\caption{
Our framework enables physically simulated robots to learn robust and generalizable interaction skills from sparse demonstrations: (top left) Learning sustained and robust dribbling from a single, brief demonstration; (top right) acquiring robust skill transitions from fragment skill demonstrations; (bottom left) generalizing book grasping to varied poses from one demonstration; and (bottom right) learning to reorientate a cube from a single grasp pose.
}
\label{fig: teaser}
\end{teaserfigure}


\maketitle

\section{Introduction}
Robot-object interaction skills are fundamental to numerous applications, ranging from character animation to robotic manipulation~\cite{xiaounified,gao2024coohoi,tennis,wang2024strategy,xu2024synchronize,fu2024humanplus,fu2024mobile}. Recent advancements in reinforcement learning from interaction demonstration (RLID) have yielded promising results in acquiring these complex skills~\cite{wang2024skillmimic,wang2023physhoi,zhang2023simulation}. By focusing on robot-object state transitions, a unified learning framework has been established, enabling the acquisition of versatile interaction skills from diverse human demonstrations efficiently. However, while current demonstration collection methods provide rich interaction examples, the captured trajectories are usually noisy and parse - only capture a limited subset of possible skill variations rather than the full spectrum of interaction patterns~\cite{menolotto2020motion,wang2024skillmimic,kim2024parahome,zhang2024core4d,jiang2023full,GRAB2020,liu2022hoi4d,fan2023arctic,fan2024hold}. Therefore, developing methods to acquire robust and generalizable interaction skills from sparse and noisy demonstrations is of particular importance.

In this work, we present a novel data augmentation and training system built upon RLID that significantly enhances its capabilities in handling imperfect demonstrations, achieving superior convergence stability, robustness to perturbations, and generalization performance. Our key insight is that despite noisy and sparse demonstrations, there exist infinite physically feasible trajectories that naturally bridge between demonstrated skills or emerge from their neighboring states, forming a continuous space of possible skill variations and transitions. Building upon this insight, we develop a comprehensive data augmentation framework to fully identify these uncaptured skill patterns. The framework consists of two core components: a Stitched Trajectory Graph (STG) that discovers potential transitions between demonstration skills, and a State Transition Field (STF) that establishes unique connections for arbitrary states within the demonstration neighborhood. To facilitate effective STF learning through RLID, we introduce an Adaptive Trajectory Sampling (ATS) strategy to ensure balanced learning of hard samples, complemented by a pre-trained history encoder for memory-dependent skill learning.

Given sparse and noisy demonstrations, our method not only acquires intended interaction skills but also achieves robust recovery capabilities from error states within the demonstration neighborhood. Furthermore, our approach masters unseen bridging transitions between demonstrated skills, enabling robust and smooth skill switching. This demonstrates the potential of our method in enriching the coverage of interaction and manipulation patterns that are typically challenging to capture during data collection.

Extensive experiments across diverse datasets, including BallPlay-M~\cite{wang2024skillmimic} and ParaHome~\cite{kim2024parahome}, demonstrate substantial improvements over state-of-the-art approaches. Our method achieves near-perfect success rates with 40-50\% improvement and enhances generalization performance by over 35\% compared to existing methods. Comprehensive ablation studies and case analyses further validate the effectiveness of each proposed component.
We encourage readers to visit our project website for video demonstrations: \textbf{https://ingrid789.github.io/SkillMimicV2/}.

\section{Related Work}

\subsection{Imitation Learning in Character Animation}
In recent years, the field of learning physics-based character skills from demonstrations has witnessed remarkable advancements \cite{DeepMimic,ase,pan2024synthesizing,dou2023c, wang2024skillmimic, wang2023physhoi, zhang2023simulation, tennis,bae2023pmp,hassan2023synthesizing,braun2023physically,sferrazza2024humanoidbench, park2019learning, xiao2025motionstreamer, liu2018learning,luo2023perpetual,tessler2023calm}. Broadly, these methods can be categorized into two types: locomotion and interaction.

\paragraph{Locomotion.} Recently, reinforcement learning \cite{kaelbling1996reinforcement} within physics-based simulation environments \cite{makoviychuk2021isaac}, guided by imitation reward functions, emerged as the mainstream approach for humanoid skill acquisition. This shift has been instrumental in both character animation \cite{DeepMimic,amp,ase} and the development of robust gaits for real-world robots \cite{he2024hover,he2024omnih2o,fu2024humanplus,he2024learning,zhang2024wococo}. \cite{DeepMimic, amp, ase} introduced classic aligned imitation reward functions and unaligned adversarial imitation reward functions \cite{ho2016generative} to learn locomotion skills. Further research has applied imitation rewards to motion-tracking \cite{luo2023perpetual} and conditional control \cite{dou2023c,tessler2023calm}. These seminal works inspired subsequent research that leverages locomotion priors for learning diverse interaction tasks~\cite{liu2024mimicking,xiaounified,ase,dou2023c,tessler2023calm,tessler2024maskedmimic,hassan2023synthesizing}, such as playing tennis \cite{tennis}, climbing ropes \cite{bae2023pmp}, and grasping~\cite{luoomnigrasp}.

\paragraph{Interaction.} A significant body of research in Human-Object Interaction (HOI) has emphasized non-physical generative approaches \cite{li2025controllable,li2023object,li2023controllable,xu2023interdiff,xu2024interdreamer,jiang2023full,Starke2020Local,Starke2021Neural,yang2025f,wang2024move,jiang2024scaling}. Despite their advantages in multimodal integration and scalability, these methods inherently lack physical authenticity and necessitate extensive training data.
Recent works have attempted to extend the success of imitation learning in locomotion to interactive skill acquisition, forming an emerging paradigm we term as Reinforcement Learning from Interaction Demonstration (RLID). Zhang et al. \cite{zhang2023simulation} introduced interaction graph for learning multi-character interactions and retargeting. Chen et al. \cite{chen2024object} developed a hierarchical policy learning framework that leverages human hand motion data to train object-centric dexterous robot manipulation. Most relevant to our work, SkillMimic \cite{wang2024skillmimic, wang2023physhoi} proposed a scalable framework for RLID with a unified interaction imitation reward, enabling the acquisition of complex basketball skills such as dribbling and shooting, while demonstrating the reusability of these learned interaction skills.

\begin{figure}[t!]
  \centering
   \includegraphics[width=1.0\linewidth]{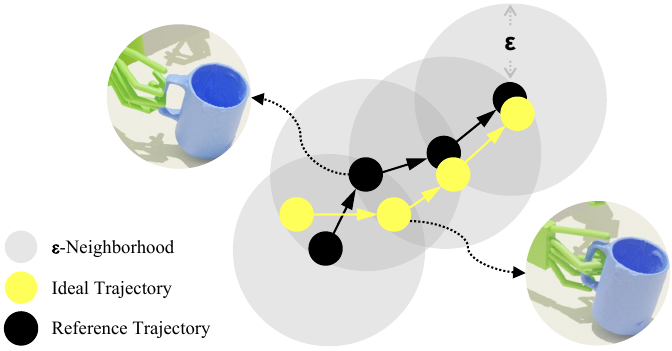}
   \caption{Given a degraded reference trajectory containing physically unreachable state transitions, perfect trajectory reconstruction becomes impossible. The goal is to learn a set of ideal trajectories that are both physically feasible and satisfy reconstruction thresholds. These ideal trajectories must exist within an $\boldsymbol{\varepsilon}$-neighborhood of the reference trajectory.}
   \label{fig: error}
   \vspace{-5mm}
\end{figure}

\subsection{Data Augmentation for Motion Data}

Data augmentation for motion capture has been a long-standing challenge in character animation. Early approaches relied on motion graphs~\cite{kovar2002motion,lee2002interactive, zhao2009achieving} to synthesize continuous animations by concatenating motion fragments. These methods typically identify similar motion frames through pose matching and resolve discontinuities via motion blending techniques \cite{kovar2003flexible,kovar2004automated}. While such approaches have achieved remarkable success in locomotion synthesis, their extension to human-object interaction (HOI) scenarios remains challenging. Motion graphs require a comprehensive motion database to enable effective transitions between motion segments. However, in HOI contexts, the introduction of manipulated objects significantly expands the interaction space, making it expensive to capture sufficient data covering all possible transition scenarios.

Recent years have witnessed the emergence of rule-based data augmentation methods for robot-object trajectories \cite{mandlekar2023mimicgen,jiang2024dexmimicgen,garrett2024skillmimicgen,gao2024efficient,pumacay2024colosseum,zhang2024diffusion}. While these approaches have shown promise in expanding manipulation datasets,
they face fundamental limitations when handling noisy demonstrations or bridging sparse motion segments in the manipulation space. In contrast, RLID~\cite{wang2023physhoi,wang2024skillmimic} has demonstrated remarkable tolerance to data noise, and generative adversarial imitation learning (GAIL) \cite{ho2016generative} with random state initialization~\cite{andrychowicz2020learning,hwangbo2019learning} has proven effective in learning generalized transitions between sparse motion segments in locomotion tasks \cite{ase}. However, the successful application of GAIL to interaction imitation remains an open challenge, since interactions require more fine-grained guidance, whereas GAN rewards tend to be coarse-grained \cite{wang2024skillmimic}.

\section{Preliminaries on RLID}

\begin{figure*}[t!]
  \centering
   \includegraphics[width=1.0\linewidth]{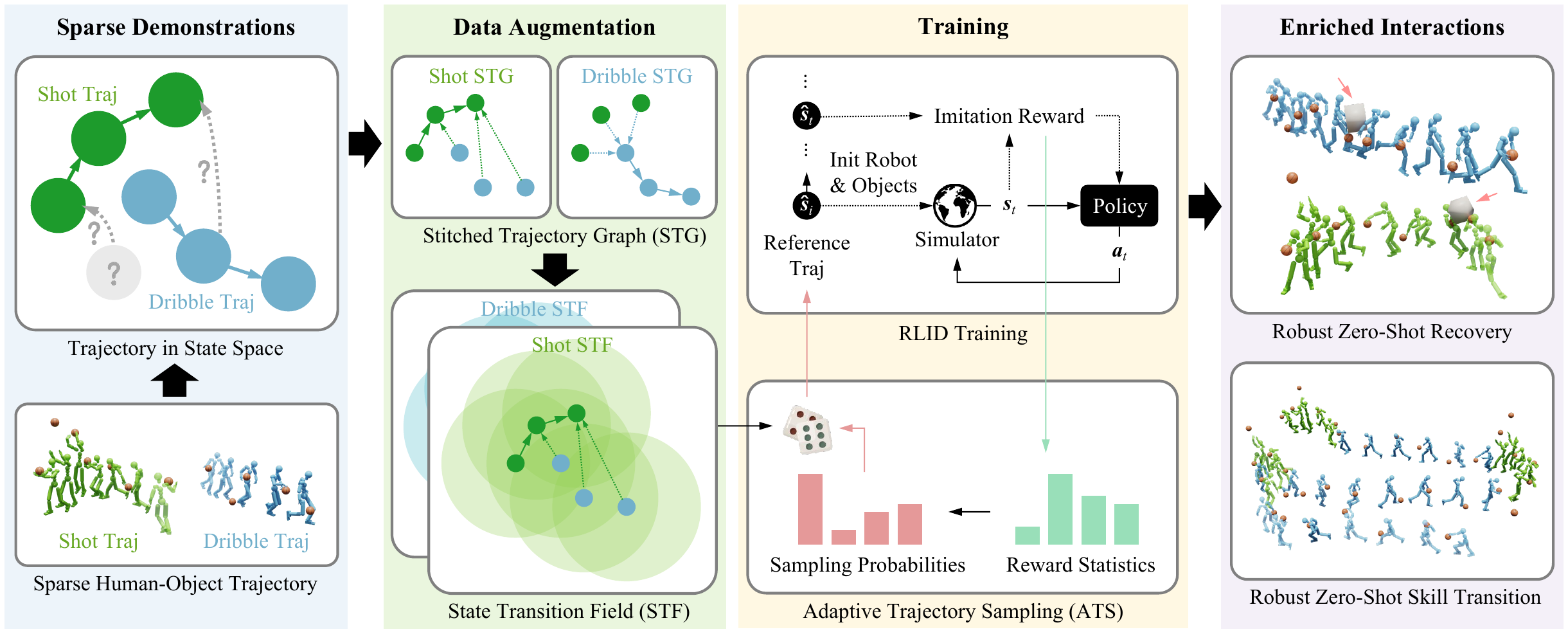}
   \caption{
   Given sparse demonstrations (e.g., two short trajectories of Shot and Dribble), there exist infinite valid but uncaptured trajectories that can either bridge between them or emerge from their neighboring states (illustrated by question marks). Our method uncovers these potential trajectories via three key steps: (1) construct a Stitched Trajectory Graph (STG) to identify possible transitions, (2) expand STG into a State Transition Field (STF) that establishes connections for arbitrary states within the demonstration neighborhood, and (3) learn a skill policy via Adaptive Trajectory Sampling (ATS) and Reinforcement Learning from Interaction Demonstrations (RLID). This enables robust skill transition and generalization far beyond the original sparse demonstrations.
   }
   \label{fig: overview}
\end{figure*}

Reinforcement Learning from Interaction Demonstration (RLID) views the learning of the manipulation task as learning underlying robot-object state transitions \cite{wang2024skillmimic}, which is typically defined by a reference trajectory $\mathcal{A}: \{\hat{\boldsymbol{s}}_0, ..., \hat{\boldsymbol{s}}_T\}$ where $T$ represents the trajectory length, $\hat{\boldsymbol{s}}_t$ represents the kinematics of both the robot and objects. 
The state transitions evolve through the interplay of a learned policy $\boldsymbol{\pi}(\boldsymbol{a}_{t}|\boldsymbol{s}_{t})$ and a deterministic physics simulator $\boldsymbol{f}(\boldsymbol{s}_{t+1}|\boldsymbol{a}_{t},\boldsymbol{s}_{t})$. The policy is parameterized as a Gaussian distribution to enable stochastic exploration, where the mean is generated by a neural network $\boldsymbol{\phi}(\boldsymbol{s}_{t})$ that maps states to actions, while maintaining a fixed variances $\boldsymbol{\Sigma}$. Formally, we have $\boldsymbol{a}_{t} \sim \mathcal{N}(\boldsymbol{\phi}(\boldsymbol{s}_{t}), \boldsymbol{\Sigma})$. We can also rewrite $\boldsymbol{s}_{t+1}$ as a stochastic variable:
\begin{equation}
\begin{aligned}
    \boldsymbol{s}_{t+1} \sim P(\cdot|\boldsymbol{\phi},\boldsymbol{s}_{t},\boldsymbol{f}).
    \label{eq: pst+1}
\end{aligned}
\end{equation}

To learn the target state transitions, a unified interaction imitation reward \cite{wang2024skillmimic} is used to measure the similarity between the generated robot-object state and the reference:
\begin{equation}
\begin{aligned}
    r_{t} = S(\boldsymbol{s}_{t+1},\hat{\boldsymbol{s}}_{t+1}) = r_{t}^{b}*r_{t}^{o}*r_{t}^{rel}*r_{t}^{cg},
    \label{eq: similarity0}
\end{aligned}
\end{equation}
which integrates four normalized sub-rewards: body states ($r_{t}^{b}$), object states ($r_{t}^{o}$), robot-object relative positions ($r_{t}^{rel}$), and contacts ($r_{t}^{cg}$). The integrated reward $r_{t}$ is bounded in [0,1], enabling consistent scaling across diverse demonstrations.
During RLID training, the robot and object states are initialized from the reference $\hat{\boldsymbol{s}}_{i}$ \cite{DeepMimic}, where $i$ is randomly sampled from $[0, T-1]$.

To handle diverse transition patterns, we adopt the conditioning mechanism from \cite{wang2024skillmimic} by introducing a condition variable $\boldsymbol{c}$ into the policy formulation $\boldsymbol{\pi}(\boldsymbol{a}_{t}|\boldsymbol{s}_{t},\boldsymbol{c})$. This variable can encode various levels of information, from high-level skill labels in basketball tasks to fine-grained target states for tracking models.

\section{Method}
\subsection{Problem Definition}
Given a noisy reference trajectory $\mathcal{A}:\{\hat{\boldsymbol{s}}_0, ..., \hat{\boldsymbol{s}}_T\}$ containing both degraded and missing states, where each state $\hat{\boldsymbol{s}}_t = [\hat{\boldsymbol{s}}_t^r, \hat{\boldsymbol{s}}_t^o]$ comprises both robot state $\hat{\boldsymbol{s}}_t^r$ and object state $\hat{\boldsymbol{s}}_t^o$. 
States are masked with $\mathcal{M}$ when missing from the trajectory.
Our goal is to learn robust interaction skills while maintaining similarity to the available reference states. Formally, we aim to learn a set $\mathcal{S}$ of feasible trajectories where each $\mathcal{A}^* \in \mathcal{S}$ maximizes the expected return $\mathcal{R}(\boldsymbol{\pi})$ while satisfies the following constraints:
\begin{equation}
\begin{aligned}
\mathcal{A}^* = \{\boldsymbol{s}^*_i, \boldsymbol{s}^*_{i+1}, ..., \boldsymbol{s}^*_T\}, \ i \in {0,...,T},
\end{aligned}
\end{equation}
\begin{equation}
\begin{aligned}
\forall t \in {i,...,T-1}: (\boldsymbol{s}^*_t, \boldsymbol{s}^*_{t+1}) \in \mathcal{F},
\label{eq: C phisics}
\end{aligned}
\end{equation}
\begin{equation}
\begin{aligned}
\forall t \in {i,...,T}: \mathcal{M}_t(|\boldsymbol{s}^*_t - \hat{\boldsymbol{s}}_t|) \leq \boldsymbol{\varepsilon}.
\label{eq: C error}
\end{aligned}
\end{equation}

Here, $\mathcal{F}$ denotes the set of physically plausible state transitions, encompassing both robot-object interaction dynamics and physical constraints. $\mathcal{M}_t \in \{0,1\}$ is a binary mask indicating whether the reference state at time $t$ is available ($\mathcal{M}_t = 1$) or missing ($\mathcal{M}_t = 0$), and $\boldsymbol{\varepsilon}$ defines the tolerance bounds for each dimension of the robot and object states. Each trajectory $\mathcal{A}^*$ can start from any time step $i$ and any state within an $\boldsymbol{\varepsilon}$-neighborhood of $\hat{\boldsymbol{s}}_i$, progressively converging towards the reference trajectory until time $T$. The set $\mathcal{S}$ represents all physically feasible trajectories in this neighborhood, characterizing the robustness and generalization of an ideal skill.

\subsection{Motivation and Method Overview}

The basic RLID method \cite{wang2024skillmimic}, which initializes states from the reference trajectory \cite{DeepMimic}, struggles with degraded data.

Unlike locomotion imitation, interaction tasks are highly sensitive to data perturbations - even a 2cm deviation between finger and objects may cause catastrophic failures. Fig.~\ref{fig: error} illustrates a typical error pattern in reference data. When data degradation exists around time step $i$, the learning of the entire state transition chain may break around time step $i$, resulting in near-zero success rates despite the policy converging well on other demonstration segments.

From Eq.~\ref{eq: C error}, we know that the target trajectory set $\mathcal{S}$ lies within the $\boldsymbol{\varepsilon}$-neighborhood of the degraded reference trajectory $\mathcal{A}$, as illustrated in Fig.~\ref{fig: error}. 
Therefore, random initialization within the entire $\boldsymbol{\varepsilon}$-neighborhood theoretically ensures complete coverage of states in $\mathcal{S}$, potentially providing better escape from local optima compared to initialization from fixed erroneous states. This insight motivates our data augmentation framework that establishes directed transitions for every state in the neighborhood, naturally forming a State Transition Field (STF). Moreover, as implied by Eq.~\ref{eq: pst+1}, increasing sampling frequency around challenging segments enhances the probability of discovering valid trajectories in $\mathcal{S}$. Similarly to \cite{won2019learning}, by allocating larger sampling weights to harder state transitions, we can better address the "chain break" problem and improve the success rate of complete trajectory execution. We term this method as Adaptive Trajectory Sampling (ATS).

Given the ability to handle noisy and incomplete data, we further augment the training data by stitching different trajectories to form a graph structure termed Stitched Trajectory Graph (STG). STF is then built upon STG to provide broader state coverage. 
For demonstrations with different condition labels, we construct separate STFs for each condition $\boldsymbol{c}$, where trajectories sampled from STF inherit the corresponding condition. During training, ATS samples reference trajectories from these STFs, allowing the policy to learn diverse state transition patterns conditioned on $\boldsymbol{c}$. Fig.~\ref{fig: overview} illustrates this process using basketball skill learning as an example. The following subsections detail these technical components.

\subsection{State Transition Field}
\label{sec: stf}

A straightforward way for neighborhood sampling is to add noise $\boldsymbol{\varepsilon}$ to the basic RLID initialization when starting from $\boldsymbol{s}_t$ \cite{liu2010sampling,ase}. However, this leads to convergence issues both theoretically and empirically. Specifically, neighborhoods of different states may have significant overlap, especially when states are close or the neighborhood range is large. In such cases, a state $\boldsymbol{s}_{\text{new}}$ may simultaneously belong to multiple reference state neighborhoods, leading to convergence challenges due to the non-unique mapping of state transitions.
Therefore, unique transition directions must be established for each neighborhood state to ensure convergence.

Moreover, when neighborhoods are large, transitions from the border to the center may be physically infeasible in a single simulation step. We resolve this by inserting masked states between distant points as potential missing data to be inpainted. These masked states contain no predefined values, serve purely as temporal buffers for bridging distant transitions and are excluded from reward computation. This essentially constructs missing data patterns that can be repaired through RLID.

The unique transition directions for all states in the neighborhood form a field of state transitions, which we term State Transition Field (STF). During training, we randomly sample trajectories from STF for RLID training, which is detailed as follows.

\subsubsection{$\boldsymbol{\varepsilon}$-Neighborhood State Initialization ($\boldsymbol{\varepsilon}$-NSI)}
\label{sec: e-NSI}
Given a reference trajectory $\mathcal{A} = \{\hat{\boldsymbol{s}}_0, ..., \hat{\boldsymbol{s}}_T\}$, we randomly select a time $i$ and sample a new state $\boldsymbol{s}_{\text{new}}$ uniformly from the $\boldsymbol{\varepsilon}$-neighborhood of the reference state $\hat{\boldsymbol{s}}_i$ as the initial state of the sampled trajectory. 

\subsubsection{Connection Rules.}
\label{sec: connection rules}
We then compute the similarity metric between $\boldsymbol{s}_{\text{new}}$ and all reference states in $\mathcal{A}$, identifying the state $\hat{\boldsymbol{s}}_j$ that exhibits maximum similarity:
\begin{equation}
\begin{aligned}
    \hat{\boldsymbol{s}}_j = \underset{s \in \mathcal{A}}{\operatorname{arg\, max}}· \ S(\boldsymbol{s}_{\text{new}}, \boldsymbol{s}).
    \label{eq: statesearch}
\end{aligned}
\end{equation}

Based on the similarity score between $\boldsymbol{s}_{\text{new}}$ and $\hat{\boldsymbol{s}}_j$, we determine the required number $N$ of masked states $\boldsymbol{s}_{\varnothing}$ to ensure feasible state transitions. The sampled trajectory is constructed as:
\begin{equation}
\begin{aligned}
\{\boldsymbol{s}_{\text{new}},\underset{N}{\underbrace{\boldsymbol{s}_{\varnothing}, ..., \boldsymbol{s}_{\varnothing}}}, \hat{\boldsymbol{s}}_j, ..., \hat{\boldsymbol{s}}_T\},
\label{eq: sampled traj}
\end{aligned}
\end{equation}
where $\boldsymbol{s}_{\text{new}}$ serves as the initialization state. The detailed computation of similarity metrics and masked node numbers is provided in the supplementary material.

\subsection{Stitched Trajectory Graph}

For sparse demonstrations, there often exist potential connections between them that were simply not captured during data collection. We can artificially construct these connections between demonstrations and use masks to indicate the missing data. Benefiting from STF's capability in handling noisy and incomplete data, these artificially introduced "noise" and "missing data" through manual stitching can be effectively repaired. This stitching approach effectively expands the coverage of the demonstration space while maintaining the inherent structure of the original demonstrations.

Consider a trajectory $\mathcal{A}$ representing skill A, and a set $\mathcal{B}$ containing all states from trajectories of other skills. We posit that all states in $\mathcal{B}$ potentially have valid transitions to skill A, even though these transitions were not captured during data collection. By stitching these potential trajectories with $\mathcal{A}$, we are essentially construct a Stitched Trajectory Graph (STG) of skill A, denoted as $\mathcal{A}^\dagger$. Specifically, for each state in $\mathcal{B}$, we employ similar connection rules as described in Sec.~\ref{sec: connection rules} to construct its path to trajectory $\mathcal{A}$. Notably, we exclude connections for states in $\mathcal{B}$ that are too distant from $\mathcal{A}$. We use the STG $\mathcal{A}^\dagger$ to replace the original reference trajectory $\mathcal{A}$ for subsequent STF data augmentation, trajectory sampling, and RLID training. Fig.~\ref{fig: overview} shows an example.

\begin{figure*}[t]
	\centering
    \subfigure[SM on Layup, 0.0\% SR]{\includegraphics[width=0.5\columnwidth]{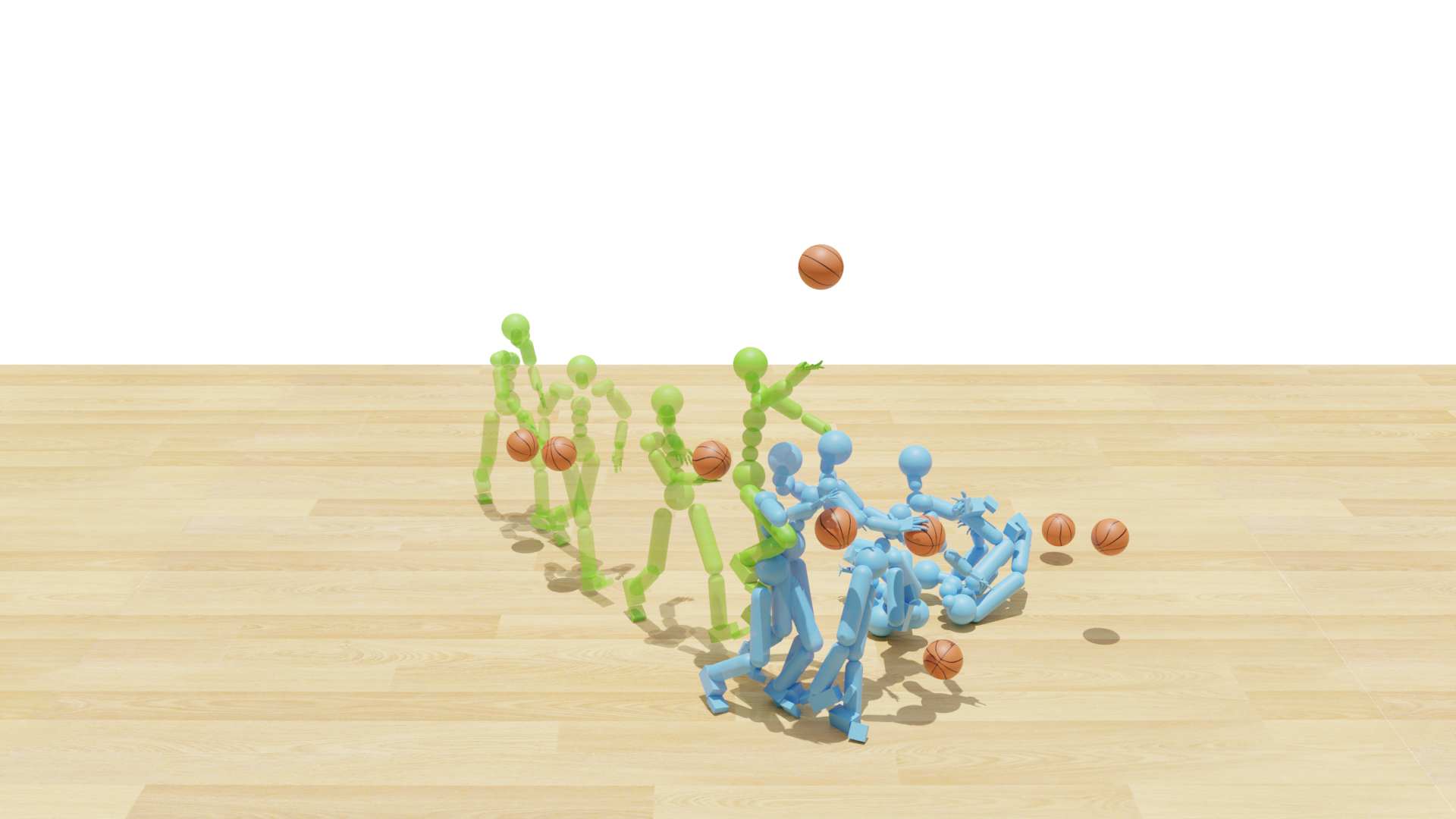}}
    \subfigure[Ours on Layup, 96.6\% SR]{\includegraphics[width=0.5\columnwidth]{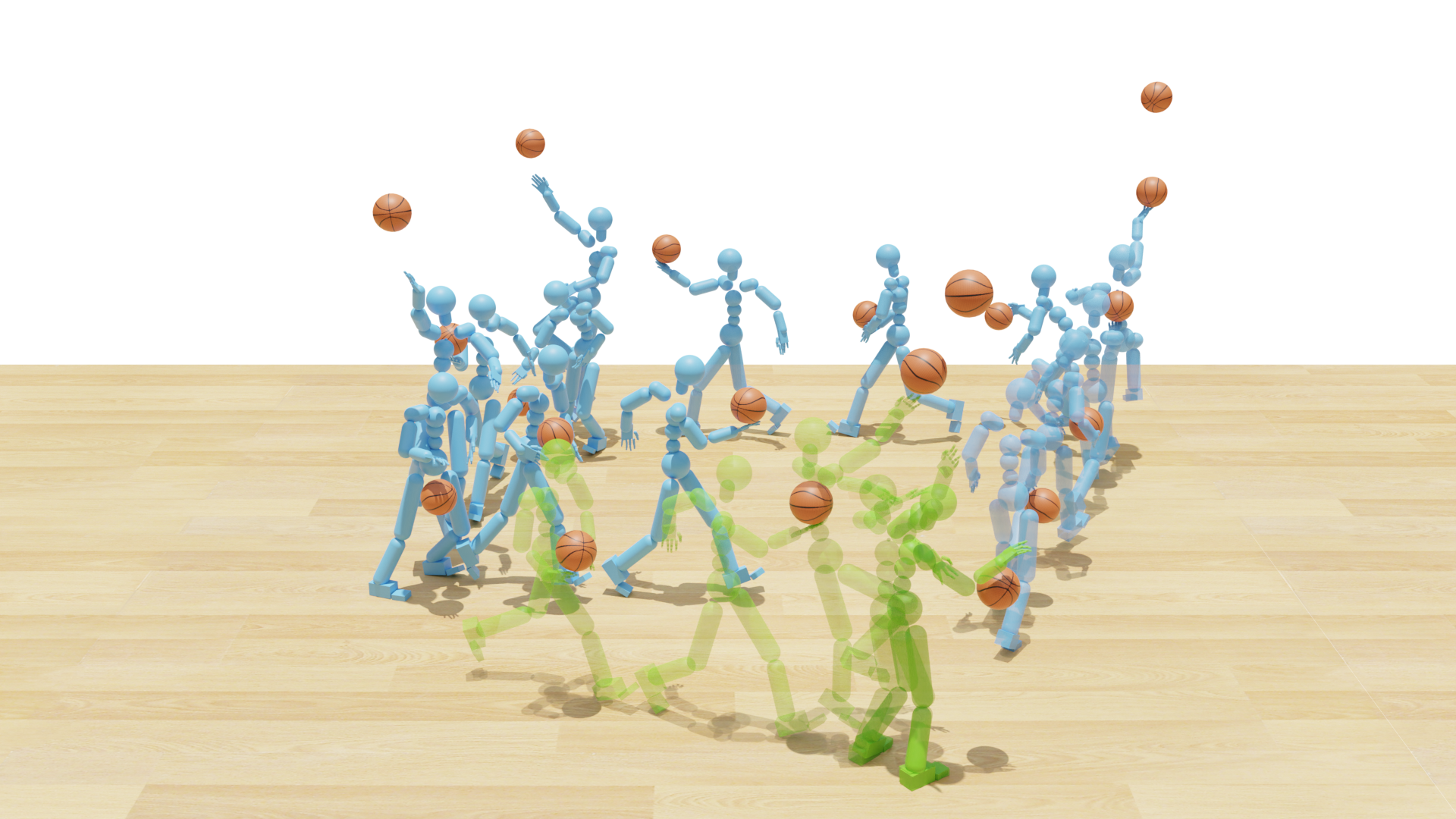}}
    \subfigure[SM on DL-DR, 1.4\% SR]{\includegraphics[width=0.5\columnwidth]{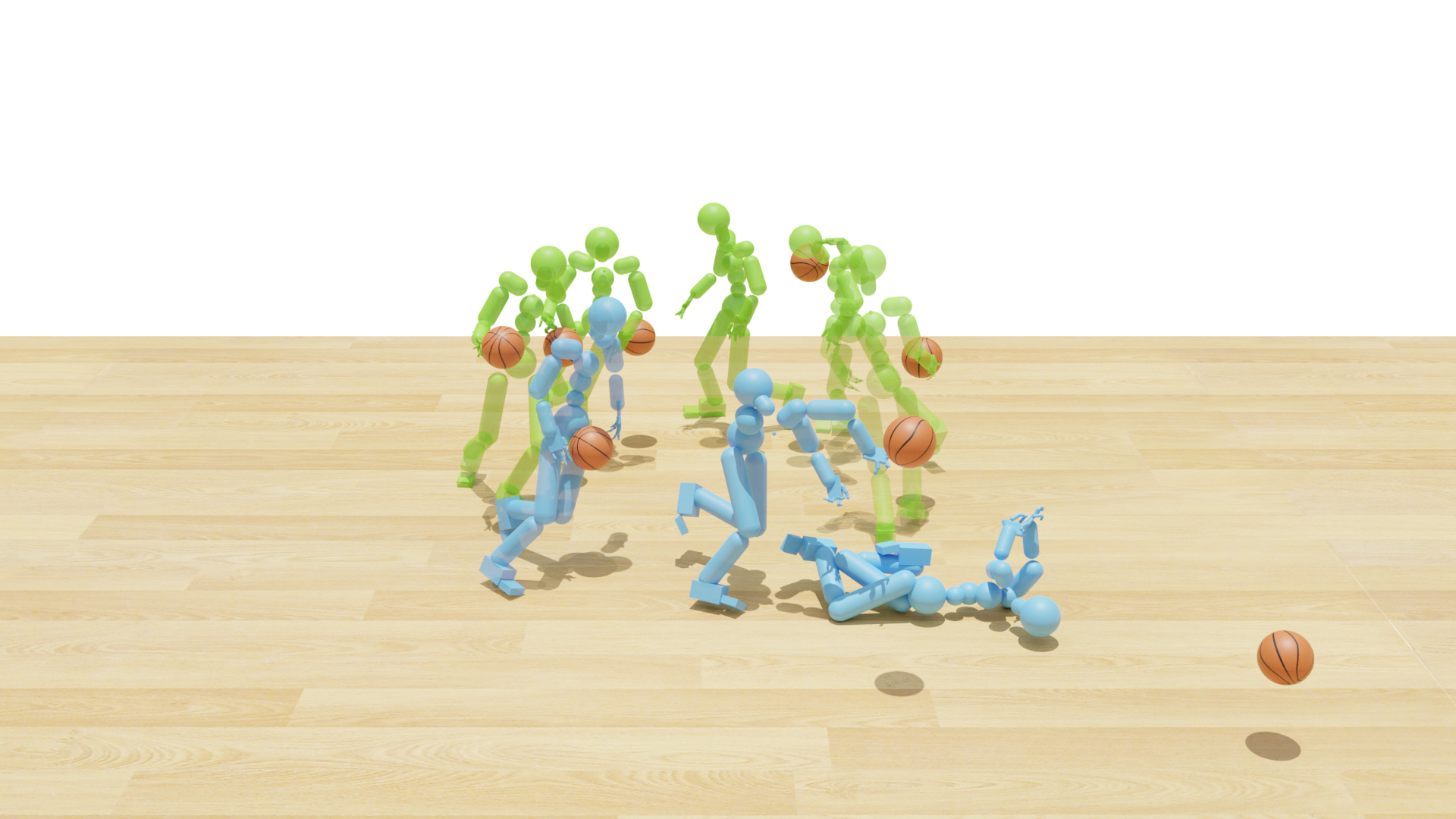}}
    \subfigure[Ours on DL-DR, 100\% SR]{\includegraphics[width=0.5\columnwidth]{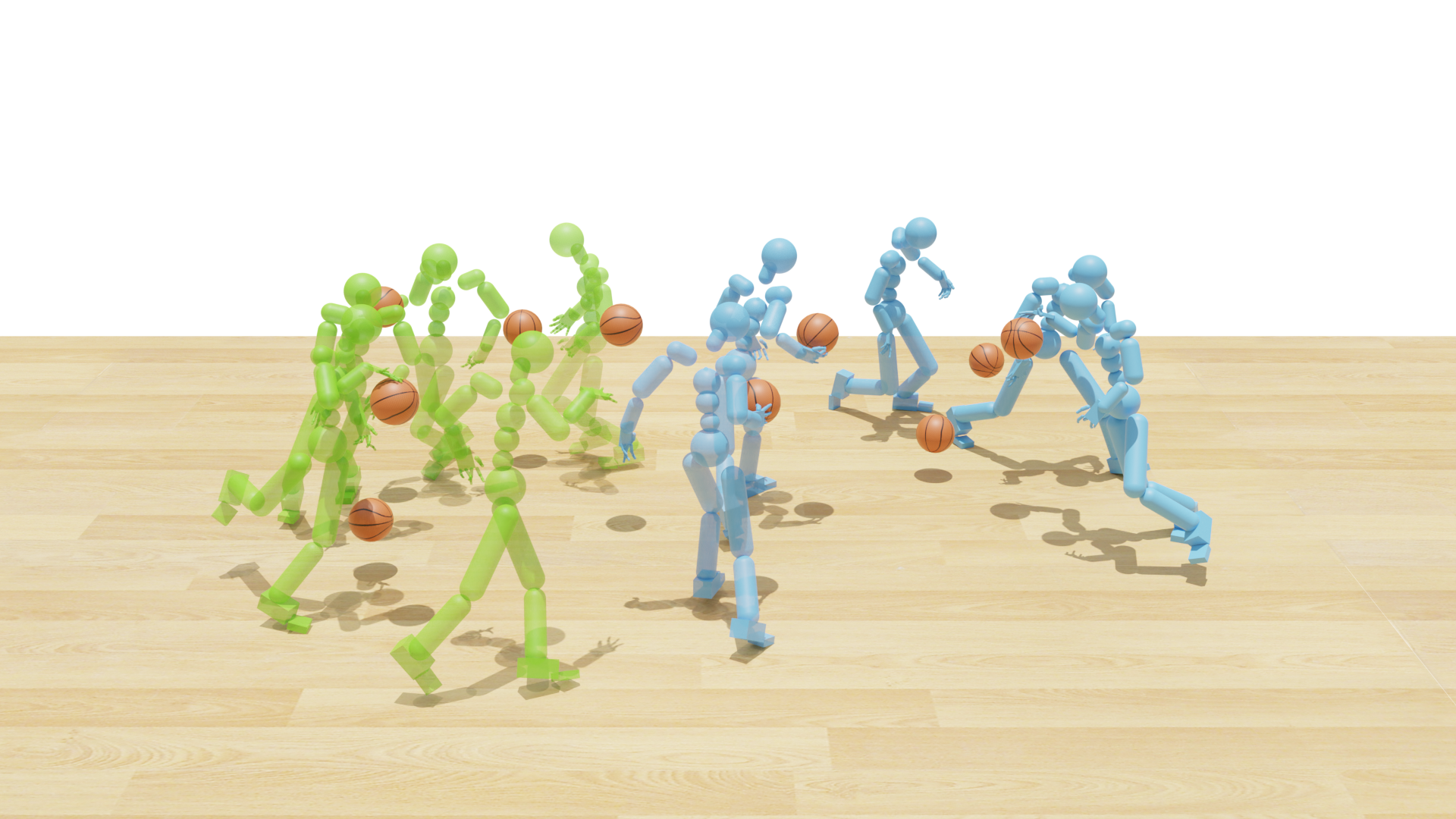}}\\
    \vspace{-2mm}
    \caption{Qualitative comparison on BallPlay-M. Blue trajectories in (a,b) indicate executions beyond the reference Layup data length. In (c,d), green and blue trajectories represent dribbling left (DL) and dribbling right (DR) respectively, demonstrating skill transition not present in the reference data.
    }
    \vspace{-2mm}
    \label{fig: ballplay}
\end{figure*}

\begin{figure*}[t]
	\centering
    \subfigure[SM on Pour-Kettle-Cup]{\includegraphics[width=1\columnwidth]{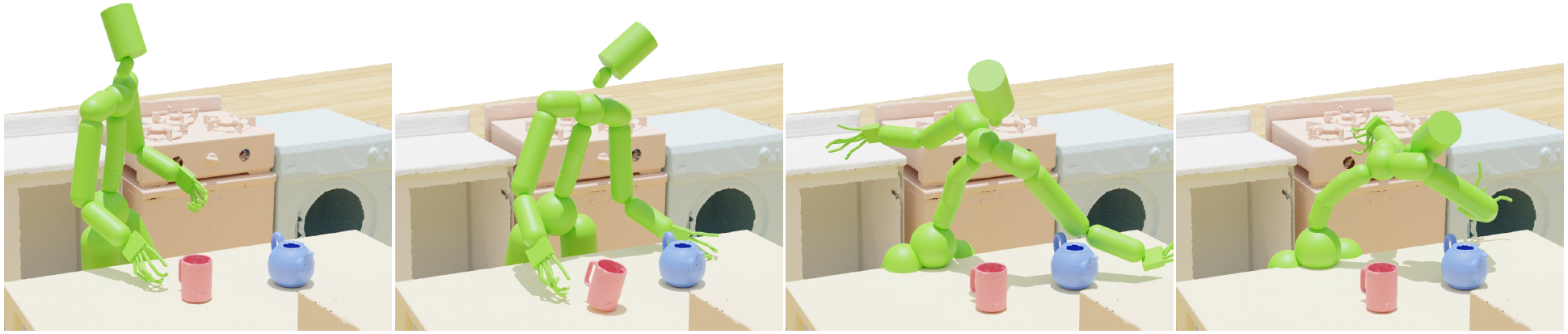}}
    \vspace{-2mm}
    \subfigure[SM on Stand-Chair]{\includegraphics[width=1\columnwidth]{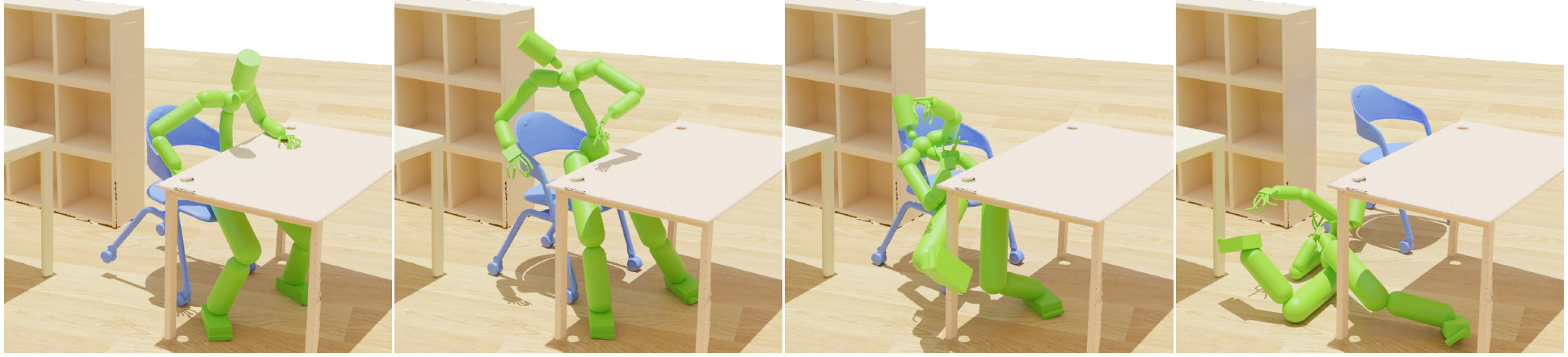}}\\
    \vspace{-2mm}
    \subfigure[Ours on Pour-Kettle-Cup]{\includegraphics[width=1\columnwidth]{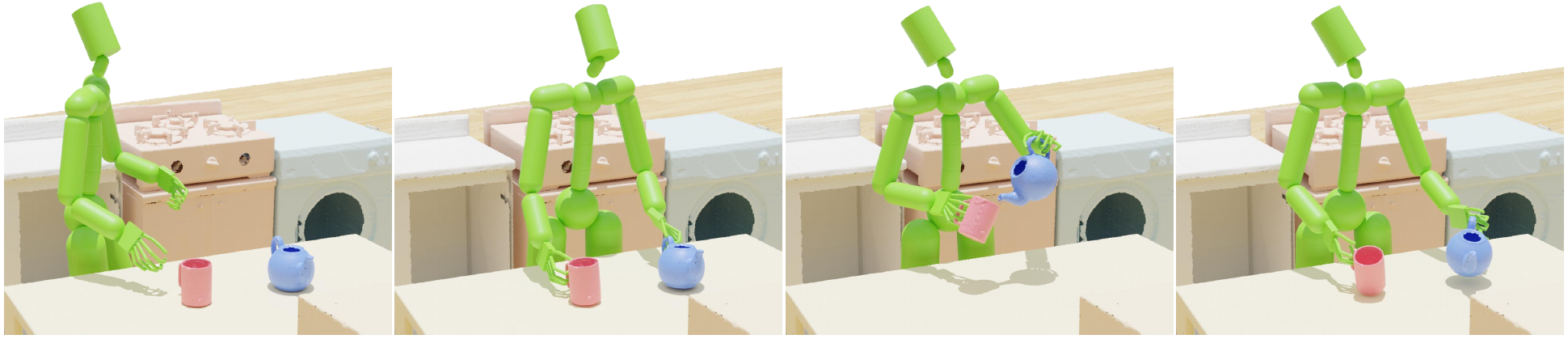}}
    \vspace{-2mm}
    \subfigure[Ours on Stand-Chair]{\includegraphics[width=1\columnwidth]{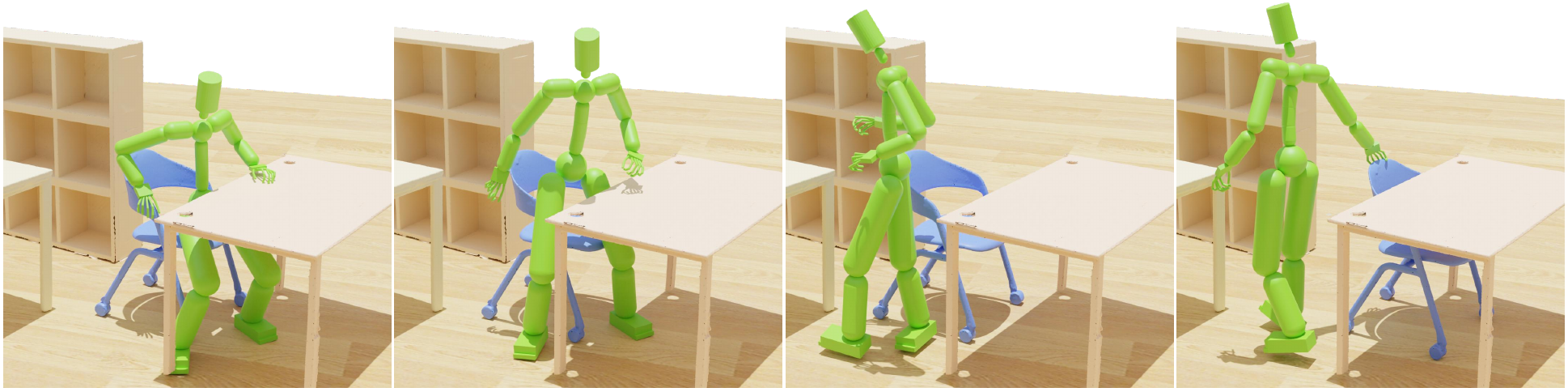}}\\
    \vspace{-2mm}
    \caption{Qualitative comparison on ParaHome. Humanoid performing (a,c) tea-pouring and teapot placement, (b,d) standing and chair-pushing sequences.
    }
    \vspace{-2mm}
    \label{fig: parahome}
\end{figure*}

\subsection{Adaptive Trajectory Sampling}

To improve performance on hard samples and address the "chain break" problem, we use Adaptive Trajectory Sampling (ATS) to adjust sampling weights based on sample difficulty.
When initialized from state $\hat{\boldsymbol{s}}_i$, the clip $\mathcal{A}_i = \{\hat{\boldsymbol{s}}_i, ..., \hat{\boldsymbol{s}}_T\}$ will be used for training. We define the sampling probability for clip $\mathcal{A}_i$ as $p_i$, formulated by:
\begin{equation}
\begin{aligned}
p_i = \frac{e^{-\lambda_s*\bar{r}_i}}{\sum_{j=0}^{T-1} e^{-\lambda_s*\bar{r}_j}},\quad
    \bar{r}_i = \frac{1}{T-i}\sum_{t=i}^{T-1} r_t,
    \label{eq: ats}
\end{aligned}
\end{equation}
where $r_t$ is defined in Eq.~\ref{eq: similarity0}, $\bar{r}_i$ is the average reward per frame, which quantifies the reconstruction quality when initializing from state $\hat{\boldsymbol{s}}_i$. $\lambda_s \in [0,\infty)$ is a coefficient that controls the trade-off between uniform sampling ($\lambda_s=0$) and difficulty-based sampling ($\lambda_s>0$). 

We now describe the complete STF sampling process integrated with ATS. Given an STF built upon STG:

\begin{itemize}
\item First, with probability $p_e$, we decide whether to sample the centroid state from external state set $\mathcal{B}$ or from the original trajectory $\mathcal{A}$. For the former case, we uniformly sample from $\mathcal{B}$. For the latter case, we select the centroid state according to the probabilities computed by ATS.

\item Once a centroid state is selected, with probability $p_n$, we either sample a starting state from its neighborhood (NSI), or use it directly as the starting state.
\end{itemize}

During multi-skill learning, skills also vary in difficulty. ATS can be similarly applied to ensure balanced learning across skill classes.

\subsection{History Encoder}

Policies lacking temporal or historical context cannot execute memory dependent behaviors, such as determining the ball-holding duration before passing. These state transitions cannot be determined solely by the current state, as similar states in the reference trajectory may lead to different transitions at different times. This ambiguity can prevent basic RLID from converging. 

While phase or temporal encoding \cite{DeepMimic} can address this issue, they require manual specification, which becomes particularly challenging when dealing with multi-skill transitions. We propose History Encoder (HE) that captures temporal dependencies in a data-driven manner, operating autonomously without manual phase specifications, enabling flexible skill transitions at arbitrary states.

Formally, given a sequence of $k$ previous states ${\boldsymbol{s}_{t-k}, ..., \boldsymbol{s}_{t-1}}$, the HE $\boldsymbol{\theta}$ generates a compact historical embedding:
\begin{equation}
\boldsymbol{h}_t = \boldsymbol{\theta}(\boldsymbol{s}_{t-k}, ..., \boldsymbol{s}_{t-1}).
\end{equation}

The policy network then takes $\boldsymbol{h}_t$ into account:
\begin{equation}
\boldsymbol{a}_t \sim \boldsymbol{\pi}(\cdot|\boldsymbol{c}, \boldsymbol{s}_t, \boldsymbol{h}_t).
\end{equation}
To ensure stable training, we pre-train HE using behavioral cloning and freeze its parameters during RLID training. This compact history embedding prevents overfitting and alleviates PPO convergence issues that might arise from high-dimensional history observations. 
Details are provided in the supplementary material.

\begin{table*}[t]
    \caption{Quantitative comparison on BallPlay-M. The neighborhood range $\boldsymbol{\varepsilon}$ for $\boldsymbol{\varepsilon}$NSR test is consistent with training settings
    }
\centering
\begingroup
\setlength{\tabcolsep}{2.5pt} 
\resizebox{\linewidth}{!}{%
\begin{tabular}{l|ccccc|c|cccc|c}
\toprule
\multicolumn{1}{c}{} & \multicolumn{6}{c}{SR$\uparrow$ (\%)  / $\boldsymbol{\varepsilon}$NSR$\uparrow$ (\%) / NR$\uparrow$ } & \multicolumn{5}{c}{TSR$\uparrow$ (\%)}
\\ 
\midrule
Method   & \textit{DF} & \textit{DL} &  \textit{DR} &  \textit{Layup}   & \textit{Shot} & \textit{Avg.} & \textit{DL-DR} & \textit{DF-DR} & \textit{DF-Shot}  &  \textit{Layup-DF} & \textit{Avg.}    \\ \midrule
DM  & 89.2 / 38.5 / 0.09 & 70.4 / 24.5 / 0.10 & 87.5 / 26.8 / 0.06  & 0.0 / 0.0 / 0.18 & 0.0 / 0.0 / 0.12 & 49.4 / 18.0 / 0.11 & 1.6 & 17.1 & 0.0 & 50.2 & 17.2 \\

DM + $\boldsymbol{\varepsilon}$-NSI  & 96.2 / 56.3 / 0.10& 76.5 / 38.7 / 0.11& 81.8 / 29.5 / 0.06& 1.0 / 0.5 / 0.20 & 0.0  / 0.0 / 0.11 & 51.1 / 25.0 / 0.12 & 2.9 & 14.0 & 0.0 & 46.1 & 15.8 \\

DM + Ours  & 83.2 / 53.2 / 0.08 & 88.3 / 55.5 / 0.09 & 92.4 / \textbf{53.4 }/ 0.10 & 78.7 / 43.7 / 0.12 & 0.6  / 0.3 / 0.06 & 68.6 / 41.2 / 0.09 & 93.4 & 87.2 & 0.0 & 71.2 & 63.0 \\

\midrule

SM  & 96.5 / 40.3 / \textbf{0.40}  & 73.0 / 27.7 / 0.49   & 96.0 / 22.7 / 0.37   & 0.0 / 0.0 / \textbf{0.64}  & 1.0 / 0.6 / 0.42   & 53.3 / 18.3 / \textbf{0.46}   & 2.1  & 26.4   & 0.8   & 31.1   & 15.1 \\

SM + $\boldsymbol{\varepsilon}$-NSI & 98.1 / \textbf{61.2} / 0.38  & 98.7 / \textbf{59.8 }/\textbf{ 0.51 }  & 97.1 / 44.9 / \textbf{0.36 }  & 23.1 / 12.1 / 0.62   & 0.0 / 0.0 / 0.39    & 63.4 / 35.6 / 0.45    & 37.2    & 77.3   & 0.0    & 49.3   & 41.0 \\

\cellcolor{gray!20}{SM + Ours}  &    \cellcolor{gray!20}   \textbf{97.7} / 60.8 / 0.37   &  \cellcolor{gray!20}  \textbf{98.5} / 59.3 / 0.42    &  \cellcolor{gray!20}  \textbf{99.1} / 47.5 / 0.34   &   \cellcolor{gray!20}  \textbf{91.5 / 44.1} / 0.57   &   \cellcolor{gray!20}  \textbf{97.9 / 34.6} / \textbf{0.46}   & \cellcolor{gray!20}  \textbf{96.9 / 49.3 } / 0.43 &  \cellcolor{gray!20}  \textbf{94.9} & \cellcolor{gray!20}  \textbf{95.7} &   \cellcolor{gray!20}  \textbf{97.2} &  \cellcolor{gray!20}  \textbf{87.4} & \cellcolor{gray!20} \textbf{93.8} \\
\midrule
\end{tabular}}
\endgroup
\label{tab: ballplay}
\end{table*}

\begin{table*}[th]
    \caption{Quantitative comparison on ParaHome. The neighborhood range $\boldsymbol{\varepsilon}$ for $\boldsymbol{\varepsilon}$NSR test is object-centric.
    }
\centering
\resizebox{\linewidth}{!}{%
\begin{tabular}{l|ccccccc|c}
\toprule
\multicolumn{1}{c}{} & \multicolumn{8}{c}{SR$\uparrow$ (\%) / $\boldsymbol{\varepsilon}$NSR$\uparrow$ (\%) / NR$\uparrow$}
\\ 
\midrule
Method & \textit{Place-Pan} & \textit{Place-Kettle} & \textit{Place-Book} & \textit{Drink-Cup} & \textit{Pour-Kettle} & \textit{Stand-Chair} & \textit{Pour--Kettle-Cup} & \textit{Avg.}\\ \midrule
SM  & 38.4 / 1.0 / 0.92  & 0.0 / 0.0 / 0.51 &  0.0 / 0.0 / 0.53 & 0.0 / 0.0 / 0.39 & 0.0 / 0.0 / 0.84 & 0.0 / 0.0 / 0.55 & 0.0 / 0.0 / 0.79 & 5.5 / 0.1 / 0.65 \\

SM + $\boldsymbol{\varepsilon}$-NSI  &  100 / 16.2 / 0.88 & 0.0 / 0.0 / 0.53 &  0.0 / 0.0 / 0.32 & 0.0 / 0.0 / 0.42 & 0.0 / 0.0  / 0.67 &  0.0 / 0.0  / 0.55 & 0.0 / 0.0  / 0.74 & 14.3 / 2.3 / 0.59 \\

SM + T  &  51.6 / 12.6 / \textbf{0.93}  & 0.0 / 0.0 / 0.30 & 100 / 12.1 / 0.85 &  99.9 / 20.3 / 0.84 & \textbf{100} / 21.5 / \textbf{0.95} & 0.0 / 0.0 / 0.63 & 48.1 / 8.0 / 0.86 & 57.1 / 15.6 / 0.77 \\

\cellcolor{gray!20}{SM + Ours}  & \cellcolor{gray!20}  \textbf{100 / 22.2} / 0.86 &  \cellcolor{gray!20}  \textbf{ 100 / 49.9 / 0.52} &  \cellcolor{gray!20}   \textbf{100 / 82.4 / 0.86 }  &  \cellcolor{gray!20}  \textbf{ 100 / 33.9 / 0.89 } &   \cellcolor{gray!20} 99.9 / \textbf{22.5} / 0.93 &  \cellcolor{gray!20}   \textbf{ 100 / 46.6 / 0.74 } &  \cellcolor{gray!20}  \textbf{ 100 / 23.1 / 0.87} &  \cellcolor{gray!20}  \textbf{ 100 / 40.1 /  0.81 } \\
\midrule
\end{tabular}}
\label{tab: parahome}
\end{table*}

\section{Experiment}

\subsection{Experimental Setup}
Our study employs Isaac Gym \cite{makoviychuk2021isaac} as the physics simulation platform. All training procedures are executed on a single NVIDIA RTX 4090 GPU, leveraging 2048 parallel environments. The PD controller and simulation operate at a frequency of 120 Hz, while the policy is sampled at 60 Hz. We use the Proximal Policy Optimization algorithm (PPO) \cite{schulman2017proximal} to optimize the policy. The simulated humanoid model replicates the kinematic tree and degrees of freedom (DoF) configurations in the demonstration dataset. Detailed hyperparameter settings are available in the Appendix.

For evaluation, we consider the following four metrics. All metrics are averaged over 10,000 random trials to ensure reliability. 

\textbf{Success Rate (SR):} the percentage of successful skill executions when initialized from the reference state of the current skill. 
For BallPlay-M~\cite{wang2024skillmimic}, we consider a skill execution successful if it can be performed continuously for 10 seconds. For ParaHome~\cite{kim2024parahome}, Success is defined as accurately reproducing the demonstrated interaction.

\textbf{Skill Transition Success Rate (TSR):} 
the percentage of successful target skill executions when initialized from other skills.

\textbf{$\boldsymbol{\varepsilon}$-Neighborhood Success Rate ($\boldsymbol{\varepsilon}$NSR):}
this metric evaluates robustness and generalization capabilities by measuring the success rate when initializing from states sampled within an $\boldsymbol{\varepsilon}$-neighborhood of the reference trajectory.

\textbf{Normalized Reward (NR):}
we compute the average reward per frame using $\text{NR} = \frac{1}{T}\sum_{t=0}^{T-1} \bar{r}_t$,
where $\bar{r}_t$ is defined in Eq.~\ref{eq: ats} and $T$ represents the length of the reference trajectory.

\subsection{Evaluation on BallPlay-M}
\label{sec: eval ballplay}

\paragraph{Dataset and Setup}
BallPlay-M \cite{wang2024skillmimic} is a human-basketball interaction dataset containing diverse basketball skills. We select 5 representative skills: Dribble-Forward (DF), Dribble-Left (DL), Dribble-Right (DR), Layup, and Shot, each represented by a 1-3 second clip. The skill labels are one-hot encoded as conditions $c$. All skills are trained using a unified policy network on a single NVIDIA RTX 4090 GPU for over 1.3 billion samples (\textasciitilde24 hours).

\paragraph{Methods}
We compare our method against two representative baselines: (1) SkillMimic (SM)~\cite{wang2024skillmimic}, a state-of-the-art RLID method, and (2) DeepMimic (DM)~\cite{DeepMimic}, a classic locomotion imitation learning approach adapted for RLID following SM's implementation \cite{wang2024skillmimic}. For fair comparison, both baselines are trained with identical setup as our method. We further augment these baselines with $\boldsymbol{\varepsilon}$-Neighborhood State Initialization ($\boldsymbol{\varepsilon}$-NSI), denoted as SM+$\boldsymbol{\varepsilon}$-NSI and DM+$\boldsymbol{\varepsilon}$-NSI respectively. Finally, we implement our full method on both baselines, denoted as SM+Ours and DM+Ours. The dimension of history embedding is 3.

\paragraph{Quantitative Analysis}
As shown in Tab.~\ref{tab: ballplay}, baseline methods achieve satisfactory performance on dribble skills (DF, DL, DR) but struggle with scoring skills (Layup, Shot). This performance gap stems from the incomplete state transition loops in reference data for scoring skills (visualized in Fig.~\ref{fig: teaser}). 
Besides, baseline methods show limited skill transition capability due to the lack of skill transition demonstrations in the reference data.

While naive $\boldsymbol{\varepsilon}$-NSI provides moderate improvements, our full method demonstrates substantial performance gains across all metrics: +45\% in average SR; +33\% in average $\boldsymbol{\varepsilon}$NSR; +84\% in average TSR. 
Although SM achieves the highest NR, demonstrating strong fitting capacity on the reference dataset, it shows unbalanced success rate and poor generalization performance. In contrast, our method not only fits the reference data well but also exhibits strong generalization and robustness to out-of-domain cases.

\begin{figure*}[t]
	\centering
    \subfigure[Comparison of Skill Success Rates at Different Training Epochs]{\includegraphics[width=1.8\columnwidth]{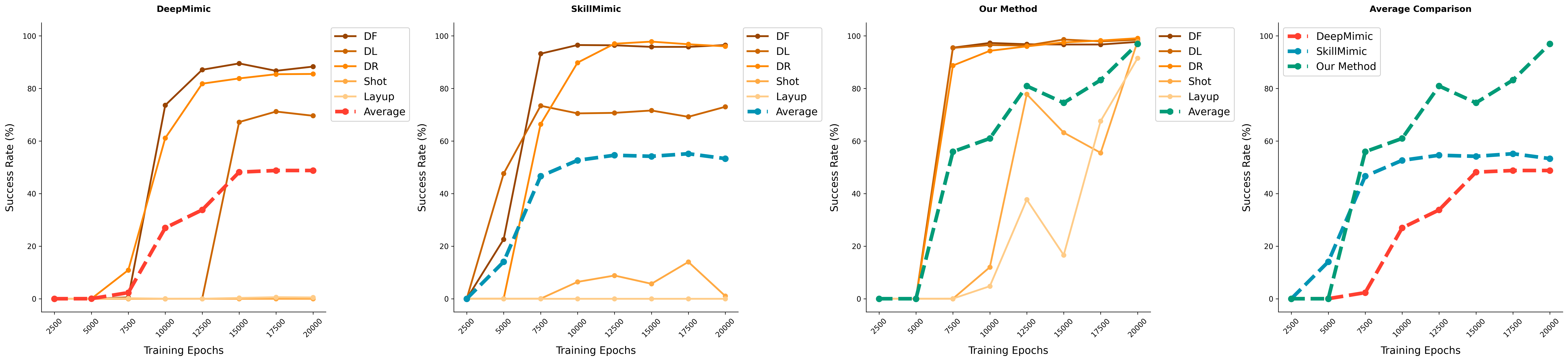}}\\
    \vspace{-3mm}
    \subfigure[Comparison of $\boldsymbol{\varepsilon}$-Neighborhood Success Rate at Different Training Epochs]{\includegraphics[width=1.8\columnwidth]{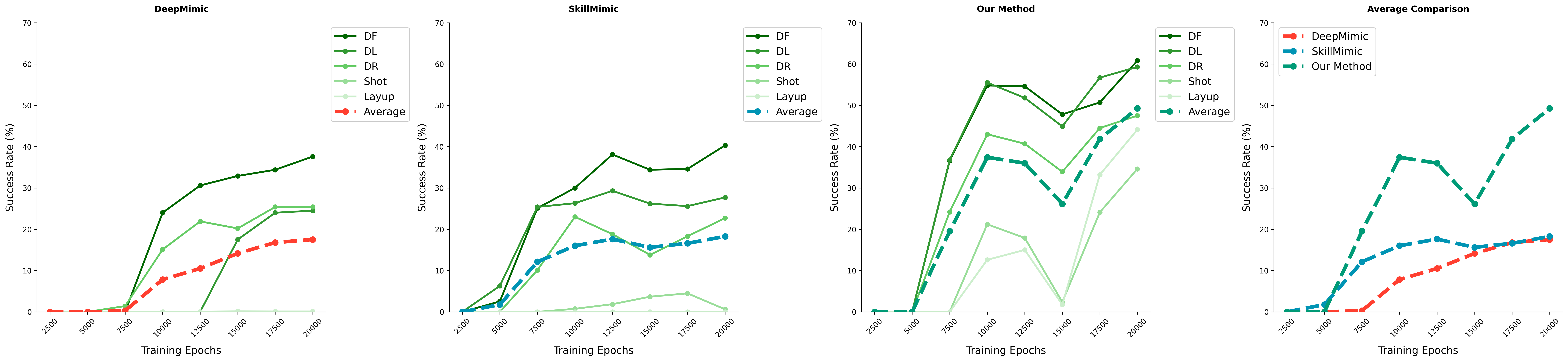}}\\
    \vspace{-3mm}
    \subfigure[Comparison of Skill Transition Success Rate at Different Training Epochs]{\includegraphics[width=1.8\columnwidth]{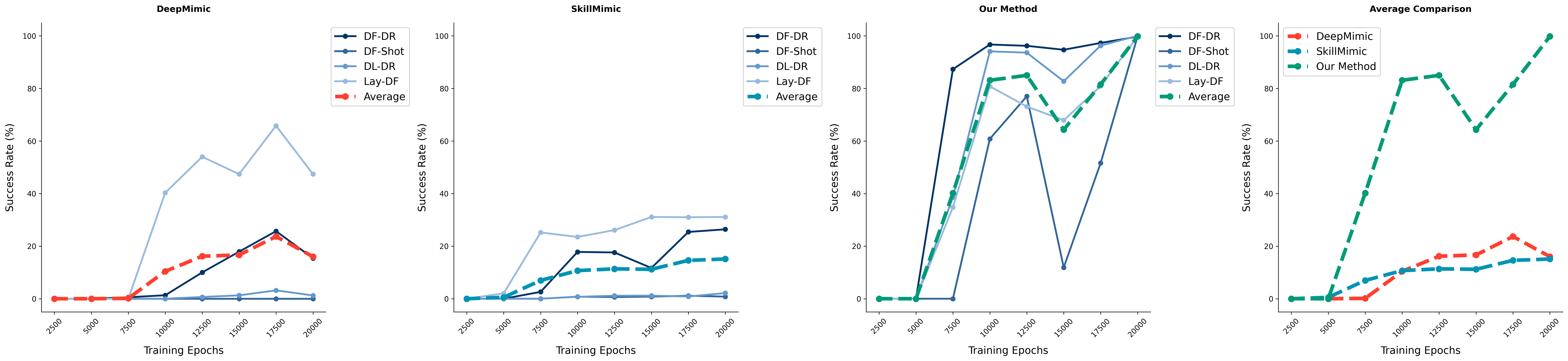}}\\
    \vspace{-3mm}
    \subfigure[Comparison of Normalized Reward at Different Training Epochs]{\includegraphics[width=1.8\columnwidth]{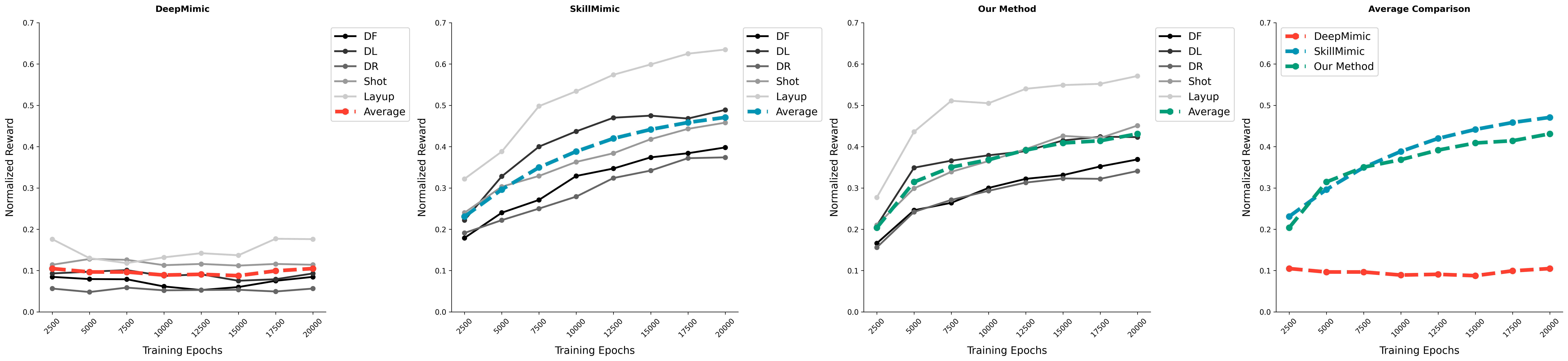}}\\
    \vspace{-3mm}
    \caption{Performance comparisons of the proposed approach against baselines across four key metrics.
    }
    \vspace{-3mm}
    \label{fig: performance at different training epochs}
\end{figure*}
\begin{figure*}[b]
    \centering
    \includegraphics[width=1.7\columnwidth]{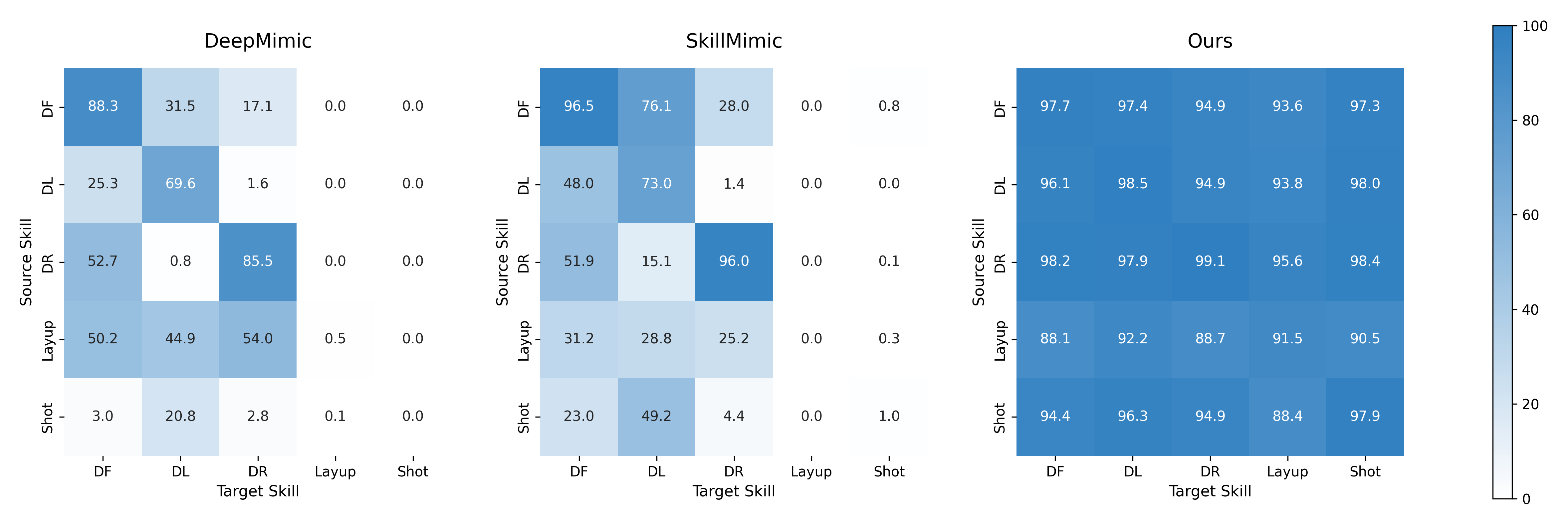}
    \vspace{-5mm}
    \caption{Comparison of skill transition success rate (\%) between five basketball skills. Our method demonstrates robust performance in achieving high success rates for transitions between arbitrary skills.
    }
    \vspace{-5mm}
    \label{fig: Detailed Skill Transition Success Rate}
\end{figure*}

In Fig.\ref{fig: performance at different training epochs}, we present a comparison of performance across different training epochs. Fig.\ref{fig: Detailed Skill Transition Success Rate} demonstrates the success rates of complete transitions among five basketball skills.

\paragraph{Qualitative Analysis}
Fig.~\ref{fig: teaser}(a) and Fig.~\ref{fig: ballplay}(b) demonstrate the superior robustness and generalization capabilities of our method, despite learning each skill from only a single noisy reference trajectory. Learning from just five sparse demonstrations, Fig.~\ref{fig: teaser}(b) showcases smooth skill transitions that never shown in the reference dataset. These visualizations, combined with the quantitative results, validate our method's capability in learning robust and generalizable interaction skills from sparse and noisy demonstrations.

\subsection{Evaluation on ParaHome}
\paragraph{Datase and Setup}
ParaHome dataset~\cite{kim2024parahome} features diverse human-object interactions in household scenarios. We evaluate on 7 representative interaction clips: Place-Pan, Place-Kettle, Place-Book, Drink-Cup, Pour-Kettle, Stand-Chair, and Pour-Kettle-Cup, each spanning 2-5 seconds. 
For each clip, we train an independent policy for around 1.0 billion samples
on a single GPU. 
To evaluate object-centric generalization, during testing, we set $\boldsymbol{\varepsilon}$ as random perturbations of object pose: -45° to 45° rotation around z-axis and up to 10cm positional offset in the xy-plane.

\paragraph{Methods}
We maintain similar method settings as in BallPlay-M experiment. However, our full approach here excludes the STG component, as trajectories involving different objects (e.g., kettle vs. chair) cannot form meaningful connections for cross-skill learning. 
We additionally condition SM with reference time $t$ as $\boldsymbol{c}$ to examine the effect of historical information, denoted as SM+T.

\paragraph{Quantitative Analysis}
The baseline method SM largely fails on these tasks due to small-scale motions and concentrated states, making action decisions challenging without historical context. Moreover, the impact of data noise is amplified in finger-level manipulation, hindering convergence. While incorporating $\boldsymbol{\varepsilon}$-NSI and T separately addresses some issues, our full method further enhances overall success rate, robustness, and convergence.

\begin{table}[t]
    \caption{Ablation study of key components on BallPlay-M.}
\centering
\resizebox{\linewidth}{!}{%
\begin{tabular}{l|cccc}
\toprule
Method  & SR$\uparrow$ (\%)& TSR$\uparrow$ (\%)& $\boldsymbol{\varepsilon}$NSR$\uparrow$ (\%)& NR$\uparrow$  \\ \midrule
SM& {53.30} & {15.11} & {18.26} & \textbf{0.47} \\
\midrule
SM + ATS& {56.39} & {22.43} & {19.74} & {0.42} \\
SM + HE& {54.06} & {20.54} & {4.33} & {0.48} \\
SM + STF& {68.67} & {35.07} & {36.96} & {0.43} \\
SM + STG& {74.74} & {71.67} & {28.91} & {0.44} \\
\midrule
SM + STG + STF& {77.12} & {73.18} & {45.08} & {0.40} \\
SM + STG + STF + ATS& {76.44} & {70.23} & {45.67} & {0.39} \\
\midrule
\cellcolor{gray!20}{SM + STG + STF + ATS + HE (Full)}& \cellcolor{gray!20}{\textbf{96.94}} & \cellcolor{gray!20}{\textbf{93.80}} & \cellcolor{gray!20}{\textbf{49.26}} & \cellcolor{gray!20}{0.43} \\

\midrule
\end{tabular}}
\vspace{-0.1cm}
\label{tab: ablation}
\end{table}

\paragraph{Qualitative Analysis}
As shown in Fig.~\ref{fig: parahome}, SM struggles to reconstruct these interactions under the compound challenges of demonstration inaccuracy and ambiguous state transitions. In contrast, our method achieves natural interaction imitation. Fig.~\ref{fig: teaser}(c) shows the generalization capabilities learned from a single demonstration.

\subsection{Ablation Study}
To assess each component's contribution, we perform comprehensive ablation studies on BallPlay-M following the experimental setup detailed in Sec.~\ref{sec: eval ballplay}. Results in Tab.~\ref{tab: ablation} reveals that each proposed component yields substantial performance gains. The synergistic integration of these components in our full method achieves the optimal performance, validating our design choices. 

For additional experimental results, including studies on data efficiency, in-hand object reorientation capabilities, locomotion skills, and comparisons with alternative methods, please refer to the supplementary material.

\section{Conclusion}
In this paper, we introduce a novel data augmentation and learning framework that fundamentally advances the learning of robust and generalizable interaction skills from sparse and noisy demonstrations. Through extensive experiments on basketball manipulations and diverse household tasks, our approach demonstrates substantial improvements over state-of-the-art methods.

While our framework shows limitations with heavily corrupted demonstrations, incorporating large-scale interaction priors (e.g., training tracking policies conditioned on target robot-object states) could address these challenges. Given our framework's unique capability to extract rich manipulation patterns from sparse noisy data, it shows promise as a fundamental building block for broader applications in both animation synthesis and robotic skill acquisition across diverse environments and tasks.


\begin{acks}
This research was supported by the Innovation and Technology Fund of HKSAR under grant number GHX/054/21GD.
\end{acks}

\bibliographystyle{ACM-Reference-Format}
\bibliography{sample-base}

\clearpage

\clearpage
\appendix

\section{Additional Experiment}

\subsection{Evaluation on Data Efficiency}
To evaluate our method's improvement in data efficiency, we conduct experiments with varying amounts of training data for a single skill. 
Specifically, we construct four datasets from BallPlay-M's pickup clips with increasing sizes: 1, 4, 10, and 40 clips respectively. For each dataset, we train policies using both SkillMimic (SM) and our method (SM+STF+ATS+HE) for approximately 3.2 billion samples. During evaluation, we place balls randomly within concentric circles of varying radii (1-5 meters) around the humanoid. 
The quantitative results in Tab.~\ref{tab: data efficiency} show our method outperforms baselines across all data scales. Even with 40 training clips, it improves generalization success rate by 13\% (reaching 96\%), demonstrating its effectiveness scales well with increased training data.

\subsection{Evaluation on Locomotion Skills}
Although our method is primarily designed for learning interaction skills, we investigate its potential benefits for locomotion skill generalization. We selected two representative locomotion skills from the BallPlay-M dataset: a single-clip Run skill and a Getup skill comprising eight diverse Getup clips. These skills were trained simultaneously using a unified policy with skill conditions.

For comparison, we also evaluated against state-of-the-art locomotion methods, specifically AMP~\cite{amp} combined with random state initialization~\cite{ase}, denoted as AMP-RSI. Other baseline settings follow those in the BallPlay-M experiment in the main paper, except that object-related terms were removed from both observations and reward functions since this experiment focuses on pure locomotion.

Tab.~\ref{tab: locomotion} presents the quantitative results, demonstrating the effectiveness of our approach in enhancing robustness and generalization performance of locomotion skills.

\subsection{Evaluation on Data Noise}
To evaluate our method's robustness against varying degrees of data degradation, we conduct experiments on BallPlay-M by introducing degradation to the reference data. We apply uniform noise sampled from [-$\sigma$, $\sigma$] on object positions, with $\sigma\in$ {10, 20, 30}mm. As shown in Tab.~\ref{tab: data noise}, our method maintains reliable performance across these challenging degradations. This is particularly noteworthy given that the original data itself contains inherent degradations.

\subsection{Evaluation on Zero-Shot In-Hand Reorientation}
While existing methods excel at grasp pose generation \cite{zhang2025graspxl,luoomnigrasp,xu2023unidexgrasp,jiang2021hand,wang2023dexgraspnet}, they typically cannot generate complex in-hand manipulations. Our method can effectively bridges this gap by augmenting discrete grasp frames into continuous manipulations. 
Specifically, to reorientate a cube to target poses, we first obtain a grasp pose using existing methods~\cite{zhang2025graspxl}. Given the geometric symmetry of the cube under 90-degree rotations, we can augment a single grasp pose into 24 valid grasp poses (6 faces × 4 orientations). Each pose is then replicated for 100 frames to create 24 trajectories, with the cube's 3D orientation serving as the condition label $c$. We train a pose-conditioned policy using our full method. During testing, given a novel cube orientation as condition, our method successfully generates natural hand manipulation sequences to achieve the desired cube orientation. Fig.~1(d) in the main paper shows an example of 90-degree cube orientation, where the intermediate manipulation process is learned from no demonstration. This application demonstrates our method's potential for both manipulation learning and data augmentation.

\begin{table}[!t]
\caption{Performance under different levels of data noise.}
\vspace{-0.1cm}
\centering
\resizebox{\linewidth}{!}{%
\begin{tabular}{l|ccc}
\toprule
\multicolumn{1}{c}{} & \multicolumn{3}{c}{ SR$\uparrow$ (\%) / $\boldsymbol{\varepsilon}$NSR$\uparrow$ (\%) / NR$\uparrow$}
\\ 
\midrule
Method  & $\sigma$ = 10 mm & $\sigma$ = 20 mm &  $\sigma$ = 30 mm \\ \midrule
SM  &  {55.8\%} / {21.9\%} / \textbf{0.45} & 56.1\% / {24.3\%} / \textbf{0.35} & {56.2\%} / {24.5\%} / \textbf{0.29}  \\

SM + Ours &   \textbf{84.9\%} / \textbf{42.5\%} / 0.38 &   \textbf{90.6\%} / \textbf{44.9\%} / 0.28 &   \textbf{90.1\%} / \textbf{53.9\%} / 0.27\\
\midrule
\end{tabular}}
\label{tab: data noise}
\end{table}

\begin{table}[t]
    \caption{Performance under different data amounts of ball pickup.
    }
\centering
\resizebox{0.7\linewidth}{!}{%
\begin{tabular}{l|cccc}
\toprule
\multicolumn{1}{c}{} & \multicolumn{4}{c}{SR$\uparrow$ (\%) with Random Ball Positions}
\\ 
\midrule
Method  & 1 Clip & 4 Clips & 10 Clips & 40 Clips \\ \midrule
SM& {0.10} & {16.26} & {32.26} & {82.84} \\
\cellcolor{gray!20}{SM + Ours}& \cellcolor{gray!20}{\textbf{0.54}} & \cellcolor{gray!20}{\textbf{46.64}} & \cellcolor{gray!20}{\textbf{85.68}} & \cellcolor{gray!20}{\textbf{96.32}} \\
\midrule
\end{tabular}}
\label{tab: data efficiency}
\end{table}

\begin{table}[t]
    \caption{Quantitative comparison on locomotion skills.
    }
\centering
\begingroup
\setlength{\tabcolsep}{3pt} 
\resizebox{\linewidth}{!}{%
\begin{tabular}{l|cc|cc}
\toprule
\multicolumn{1}{c}{} & \multicolumn{2}{c}{ SR$\uparrow$ (\%) / $\boldsymbol{\varepsilon}$NSR$\uparrow$ (\%) / NR $\uparrow$ }& \multicolumn{2}{c}{ TSR$\uparrow$ (\%)}
\\ 
\midrule
Method  & Getup & Run & Getup-Run & Run-Getup \\ \midrule
AMP + RSI& {\textbf{99.3} / \textbf{98.6} / 0.01 } & {93.3 / 80.5 / 0.66 } & {37.9} & {99.7} \\
\hline
DM& {24.4 / 24.9 / 0.64} & {46.5 / 22.2 / 0.73 } & {0.2} & {22.4} \\
DM + $\boldsymbol{\varepsilon}$-NSI  & {96.9 / 97.9 / 0.47} & { 93.2 / 84.9 / 0.65} &  {62.6} &  {5.7}  \\
DM + Ours& 98.5 / 98.2 / 0.54 & {97.1 / 81.5 / 0.64} & {67.2} & {96.2} \\
\hline
SM& {69.2 / 66.0 / \textbf{0.80} } & {74.0 / 36.4 / \textbf{0.78} } & {10.5} & {44.9} \\
SM + $\boldsymbol{\varepsilon}$-NSI & { 97.8 / 97.3 / 0.66} &  99.4 / \textbf{91.8} / 0.77 &  93.4 &  {12.7} \\
\cellcolor{gray!20}{SM + Ours}& \cellcolor{gray!20} 99.1 / 98.0 / 0.64  & \cellcolor{gray!20} \textbf{99.9} / 91.0 / 0.71 & \cellcolor{gray!20} {\textbf{97.9}} & \cellcolor{gray!20} \textbf{100.0} \\
\midrule
\end{tabular}}
\endgroup
\label{tab: locomotion}
\end{table}

\begin{table}[!t]
    \caption{More Comparisons.}
\vspace{-0.1cm}
\centering
\resizebox{0.7\linewidth}{!}{%
\begin{tabular}{l|cccc}
    \toprule
    Method  & SR$\uparrow$ (\%)& TSR$\uparrow$ (\%)& $\boldsymbol{\varepsilon}$NSR$\uparrow$ (\%)& NR$\uparrow$  \\ \midrule
    SM& {53.30} & {15.11} & {18.26} & {0.47} \\
    \midrule
    SM + HE& {54.06} & {20.54} & {4.33} & \textbf{0.48} \\
    SM + HS& {0.0} & {0.0} & {0.0} & {0.21} \\
    \midrule
    SM + STF& \textbf{68.67} & \textbf{35.07} & \textbf{36.96} & {0.43} \\
    SM + IAE& {53.31} & {17.18} & {17.09} & {0.45} \\
    \midrule
\end{tabular}}
\vspace{-0.1cm}
\label{tab: more comparison}
\end{table}

\subsection{More Comparisons}
\begin{figure}
    \centering
    \includegraphics[width=\columnwidth]{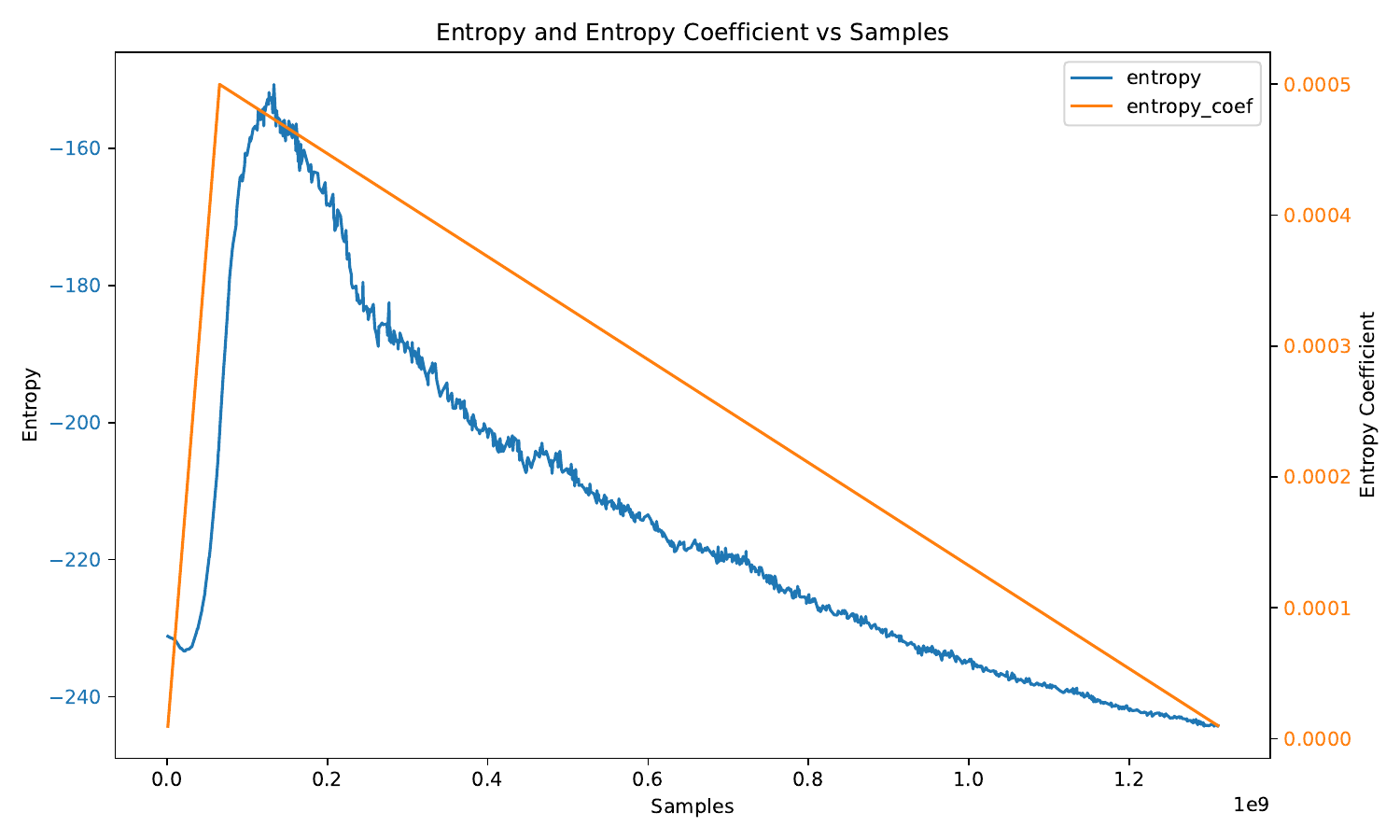}
    \caption{Increasing and annealing the exploration rate in vanilla PPO}
    \label{fig:entropy_curves}
\end{figure}

We perform additional experiments to provide thorough validations for the proposed components, comparing our designed methods against simpler, baseline alternatives.
\paragraph{Exploration Strategy:} We compare our State Transition Field (STF) method with a more straightforward alternative: increasing and annealing the exploration rate (IAE) by scheduling the entropy coefficient in vanilla PPO. Specifically, we implement a linear warm-up scheduler to dynamically adjust the entropy coefficient, initially rising to a peak (entropy coefficient $=5e^{-4}$) in the first 1000 epochs, then annealing down to a minimal level ($1e^{-5}$) by epoch 20000. Fig.~\ref{fig:entropy_curves} illustrates the entropy and entropy coefficient trajectories during training.  This entropy-based scheduling increases the corresponding sigma from approximately $0.055$ (fixed sigma baseline) to $0.093$, and subsequently anneals it back down to around $0.051$. However, this entropy-based method (SM+IAE) provides virtually no improvement over the baseline SM (see Tab.~\ref{tab: more comparison}) and significantly underperforms compared to STF.
Our objective is to learn a policy that knows how to act across a neighborhood of states. However, adjusting action variance does not directly yield such neighborhood coverage. Besides, if the variance is too large, it hinders policy learning and may even prevent convergence. Our proposed STF explicitly structured neighborhood exploration, significantly surpasses this baseline on all metrics.

\paragraph{History Representation:} Additionally, we examine the proposed History Encoder (HE) against a naïve baseline of directly concatenating 60 consecutive historical states to the policy inputs (SM+HS). Although concatenating a large window of states provides rich context, it severely hampers PPO convergence due to high-dimensional observation spaces, resulting in a collapse across all metrics, as demonstrated in Tab.~\ref{tab: more comparison}. In contrast, our HE approach, which compresses temporal history into a compact embedding, maintains stable and effective training.

\section{Technical Details}

\subsection{Observation}
The state $\boldsymbol{s}_{t} = \{\boldsymbol{o}^{sbj}_{t},\boldsymbol{o}^{f}_{t},\boldsymbol{o}^{obj}_{t}\}$ observed by the policy consists of the following components:

\begin{itemize}
    \item Humanoid observation $\boldsymbol{o}^{sbj}_{t}$:
    \begin{itemize}
        \item Global root height
        \item Body position and rotation in local coordinates
        \item Body position velocity and angular velocity
    \end{itemize}
    
    \item Contact observation $\boldsymbol{o}^{f}_{t}$:
    \begin{itemize}
        \item Net contact forces at fingertips
    \end{itemize}
    
    \item Object observation $\boldsymbol{o}^{obj}_{t}$:
    \begin{itemize}
        \item Position and rotation in local coordinates
        \item Linear and angular velocities
    \end{itemize}
\end{itemize}

All coordinates are transformed into the humanoid's root local coordinate system to enhance generalization.

\subsection{Policy}
The policy output is parameterized as a Gaussian distribution:
\begin{equation}
\begin{aligned}
    \boldsymbol{a}_{t} \sim \mathcal{N}(\boldsymbol{\phi}_{\boldsymbol{\pi}}(\boldsymbol{s}_{t},\boldsymbol{h}_{t},\boldsymbol{c}), \boldsymbol{\Sigma}_{\boldsymbol{\pi}}),
\end{aligned}
\end{equation}
where $\boldsymbol{\phi}_{\boldsymbol{\pi}}$ is a three-layer MLP (1024-512-512 units, ReLU activations) that maps state $\boldsymbol{s}_{t}$, history embedding $\boldsymbol{h}_{t}$, and skill condition $\boldsymbol{c}$ to action means. The variances $\boldsymbol{\Sigma}_{\boldsymbol{\pi}}$ are set to 0.055 during training for exploration and 0 during testing for stability. The action $\boldsymbol{a}_{t}$ represents target joint rotations, which are processed by a PD controller to generate joint torques.

\subsection{Reward Function}
Following SkillMimic~\cite{wang2024skillmimic}, we use a unified imitation reward for RLID training. The imitation reward combines four components:
\begin{equation}
\begin{aligned}
    r_{t} = S(\boldsymbol{s}_{t+1},\hat{\boldsymbol{s}}_{t+1}) = r_{t}^{b}*r_{t}^{o}*r_{t}^{rel}*r_{t}^{cg},
\end{aligned}
\end{equation}
where $r_{t}^{b}$, $r_{t}^{o}$, $r_{t}^{rel}$, and $r_{t}^{cg}$ represent body motion, object motion, relative motion, and contact graph rewards respectively. All reward weights are listed in Tab.~\ref{tab: sm reward weight}.

\begin{itemize}
\item \textbf{Body Motion Term:}
\begin{equation}
\begin{aligned}
    r_{t}^{b} = r_{t}^{p}*r_{t}^{r}*r_{t}^{pv}*r_{t}^{rv},
\end{aligned}
\end{equation}
Each sub-term follows:
\begin{equation}
    r_{t}^{\alpha} = e^{-\lambda^{\alpha} *\text{MSE}{(\boldsymbol{s}_{t+1}^{\alpha},\hat{\boldsymbol{s}}_{t+1}^{\alpha})}},
\end{equation}
where $\alpha \in \{p,r,pv,rv\}$ represents position, rotation, position velocity, and rotation velocity respectively. $\hat{\boldsymbol{s}}_{t+1}^{\alpha}$ and $\boldsymbol{s}_{t+1}^{\alpha}$ denote reference and simulated states.

\item \textbf{Object Motion Term:}
\begin{equation}
\begin{aligned}
    r_{t}^{o} = r_{t}^{op}*r_{t}^{or}*r_{t}^{opv}*r_{t}^{orv},
\end{aligned}
\end{equation}
with sub-terms following the same formulation as body motion rewards.

\item \textbf{Relative Motion Term:}
\begin{equation}
\begin{aligned}
    r_{t}^{rel} = e^{-\lambda^{rel} *\text{MSE}{(\boldsymbol{s}_{t+1}^{rel},\hat{\boldsymbol{s}}_{t+1}^{rel})}},
\end{aligned}
\end{equation}

\item \textbf{Contact Graph Term:}
\begin{equation}
\begin{aligned}
    r_{t}^{cg} = e^{-\sum_{j=1}^{J}\boldsymbol{\lambda}^{cg}[j] *\boldsymbol{e}_{t+1}^{cg}[j]},
\end{aligned}
\end{equation}
where $\boldsymbol{e}_{t+1}^{cg} = |\boldsymbol{s}_{t+1}^{cg}-\hat{\boldsymbol{s}}_{t+1}^{cg}|$ represents the contact error between simulated and reference states. $J$ is the number of contact pairs. For experiment on BallPlay-M, $\boldsymbol{s}^{cg}$ contains three contact pairs: ball-hands contact, ball-body contact, and body-hands contact. Due to Isaac Gym's limitations in detecting complex contact pairs \cite{makoviychuk2021isaac}, we determine contacts based on contact forces. For experiment on ParaHome~\cite{kim2024parahome}, the contact graph reward is disabled (i.e., $r_{t}^{cg}=1$).
\end{itemize}

For DM, we follow the implementation of SM, with the only modification being the adoption of DM-style additive reward:
\begin{equation}
\begin{aligned}
    r_{t} = r_{t}^{p} + r_{t}^{r} + r_{t}^{rv} + r_{t}^{op},
\end{aligned}
\end{equation}
with reward weights listed in Tab.~\ref{tab: dm reward weight}.

\subsection{Connection Rules}
\label{sec: connection rules supp}
Given any two states $\boldsymbol{s}_{A}$ and $\boldsymbol{s}_{B}$, their kinematic similarity is evaluated using a modified similarity metric that excludes contact information:
\begin{equation}
\begin{aligned}
    S_{k}(\boldsymbol{s}_{A},\boldsymbol{s}_{B}) = r^{b}*r^{o}*r^{rel},
    \label{eq:state distance}
\end{aligned}
\end{equation}
where $r^{b}$, $r^{o}$, and $r^{rel}$ are defined identically to those in the reward function. Let $\beta = S_{k}(\boldsymbol{s}_{A},\boldsymbol{s}_{B})$ denote the computed similarity score, the connection from $\boldsymbol{s}_{A}$ to $\boldsymbol{s}_{B}$ is established according to the following criteria, where $\tau$ represents a predetermined similarity threshold:

\begin{itemize}
\item When $\beta > \tau$, we introduce intermediate masked states between $\boldsymbol{s}_{A}$ and $\boldsymbol{s}_{B}$. The number of masked states is determined by:
\begin{equation}
\begin{aligned}
    N = \min(-\lfloor\log_{10}(\beta)\rfloor, N_{max}),
\end{aligned}
\end{equation}
where $N_{max}$ denotes the maximum allowable number of masked states.

\item When $\beta < \tau$, the connection is deemed invalid and subsequently discarded from consideration.
\end{itemize}
For BallPlay-M~\cite{wang2024skillmimic}, when constructing the stitched trajectory graph (STG), we apply coordinate transformation to align the stitched state pairs. Specifically, for a state $\boldsymbol{s}_A$, we first transform its root position by aligning its (x,y) coordinates with the reference state $\boldsymbol{s}_B$'s root before computing their similarity. The transformed $\boldsymbol{s}_A$ is then evaluated against connection criteria to determine whether it should be added to the STG.

\begin{algorithm}[htbp]
\caption{Online Motion Data Augmentation}
\label{alg:data_augmentation}
\KwIn{Dataset $\mathcal{D}$, probabilities $p_1, p_2$, neighborhood radius $\epsilon$}
\KwOut{Augmented motion sequence $\tilde{\mathbf{m}}$}

Sample reference skill motion $\mathbf{m} = \{\hat{\boldsymbol{s}}_0, ..., \hat{\boldsymbol{s}}_T\}$ from $\mathcal{D}$\;
\If{$\text{Bernoulli}(p_1)$}{
    Sample reference skill motion $\mathbf{n} \neq \mathbf{m}$ from $\mathcal{D}$\;
    Sample initial state $\hat{\boldsymbol{s}}^{*}$ from $\mathbf{n}$\;
}
\Else{
    Sample $k \in [0, T]$ according to ATS\;
    $\hat{\boldsymbol{s}}^{*} \leftarrow \hat{\boldsymbol{s}}_{k}$\;
}

\If{$\text{Bernoulli}(p_2)$}{
    Sample neighborhood states $\hat{\boldsymbol{s}}_{nb} \sim \mathcal{N}(\hat{\boldsymbol{s}}^{*}, \epsilon)$\;
    $\hat{\boldsymbol{s}}^{*} \leftarrow \hat{\boldsymbol{s}}_{nb}$\;
}

\For{$i \in [0, T]$}{
Compute similarity scores $d_i = \text{Dist}(\hat{\boldsymbol{s}}^{*}, \hat{\boldsymbol{s}}_i)$ 
(Eq.~\ref{eq:state distance})\;
}
Find closest state index $j = \arg\min_i d_i$\;
Calculate mask length $N$ according to $d_j$ following Sec.~\ref{sec: connection rules supp}\;

Return augmented sequence $\tilde{\mathbf{m}} = \{\hat{\boldsymbol{s}}^{*},\underset{N}{\underbrace{\boldsymbol{s}_{\varnothing}, ..., \boldsymbol{s}_{\varnothing}}}, \hat{\boldsymbol{s}}_j, ..., \hat{\boldsymbol{s}}_T\}$\;

\end{algorithm}

\begin{table}[t]
\centering  
\caption{Hyperparameters for policy training.}
\label{tab: params_training}
\begin{tabular}{|l|c|}
    \toprule
    {\bf Parameter} & {\bf Value}  \\ 
    \hline
    $\mathrm{dim}(\boldsymbol{c})$ Skill Embedding Dimension &  $64$  \\ \hline
    $\Sigma_{\boldsymbol{\pi}}$ Action Distribution Variance &  $0.055$  \\ \hline
    Samples Per Update Iteration &  $65536$  \\ \hline
    Policy/Value Function Minibatch Size &  $16384$  \\ \hline
    $\gamma$ Discount &  $0.99$  \\ \hline
    Adam Stepsize & $2 \times 10^{-5}$ \\ \hline
    GAE($\lambda$) &  $0.95$  \\ \hline
    TD($\lambda$) &  $0.95$  \\ \hline
    PPO Clip Threshold &  $0.2$  \\ \hline
    $T$ Episode Length &  $60$  \\ \hline
    $\mu$ Dimension of history embedding &  $3$  \\ \hline
    $k$ History horizon length &  $60$  \\ 
    \bottomrule
\end{tabular}
\end{table}

\begin{table}[t]
{ \centering  
\caption{Hyperparameters for Data Augmentation. Note that $N_{max}$ represents the max allowable number of masked states; $\tau$ means the state similarity threshold; $p_e$ is the probability of sampling from external reference states; $p_n$ is the probability of sampling from external states; $\lambda_s$ is the Adaptive Trajectory Sampling (ATS) weight; $\lambda_c$ is the inter-class ATS weight.}
\label{tab: params_augmentation}
\resizebox{1\linewidth}{!}{%
\begin{tabular}{|l|c|c|c|}
\hline
{\bf Parameter} & {\bf BallPlay-M} & {\bf Locomotion} & {\bf ParaHome}  \\ \hline
    $N_{max}$ &  $10$ &  $10$ &  $10$ \\ \hline
    $\tau$ &  $1 \times 10^{-10}$&  $1 \times 10^{-10}$ &  $1 \times 10^{-10}$  \\ \hline
    $p_e$ &  $0.1$ &  $0.1$ &  $-$ \\ \hline
    $p_n$ &  $0.1$ &  $0.1$ &  $0.1$\\ \hline
    $\lambda_s$ & 10 & 10 & 10 \\ \hline
    $\lambda_c$ & 5 & 5 & 5 \\ \hline
    $\boldsymbol{\varepsilon}_{rootpos}$ &  $0.1$ & 0.1 for Run. 1.0 for Getup &  $0.1$ \\ \hline
    $\boldsymbol{\varepsilon}_{rootvel}$ &  $0.1$ & 0.1 for Run. 1.0 for Getup &  $0.1$ \\ \hline
    $\boldsymbol{\varepsilon}_{rootrot}$ & $0.1$  & $0.1$ & $0.1$ \\ \hline
    $\boldsymbol{\varepsilon}_{rootrotvel}$ & $0.1$  & $0.1$ & $0.1$ \\ \hline
    $\boldsymbol{\varepsilon}_{dof}$ & $0.1$  & $0.1$ & $0.1$ \\ \hline
    $\boldsymbol{\varepsilon}_{dofvel}$ & $0.1$  & $0.1$ & $0.1$ \\ \hline
    $\boldsymbol{\varepsilon}_{objpos}$ & $0.1$  & $0.1$ & $0.1$ \\ \hline
    $\boldsymbol{\varepsilon}_{objposvel}$ & $0.1$  & $0.1$ & $0.1$ \\ \hline
    $\boldsymbol{\varepsilon}_{objrot}$ & $0.1$  & $0.1$ & $0.1$ \\ \hline
    $\boldsymbol{\varepsilon}_{objrotvel}$ & $0.1$  & $0.1$ & $0.1$\\ \hline
\end{tabular}
}
}
\end{table}

\begin{table}[t]
\centering
\caption{Reward weights of SM in different tasks.}
\resizebox{\linewidth}{!}{%
\begin{tabular}{|l|c|c|c|}
\hline
\textbf{Parameter} & \textbf{BallPlay-M} & \textbf{Locomotion} & \textbf{ParaHome} \\ \hline
$\lambda^{p}$ Position & $20$ & $20$ & $20$ \\ \hline
$\lambda^{r}$ Rotation & $20$ & $20$ & $20$ \\ \hline
$\lambda^{pv}$ Velocity & $0$ & $0$ & $0$ \\ \hline
$\lambda^{rv}$ Rotation Velocity & $0$ & $0$ & $0$ \\ \hline
$\lambda^{op}$ Object Position & $1$ & $-$ & $1$ \\ \hline
$\lambda^{or}$ Object Rotation & $0$ & $-$ & $0$ \\ \hline
$\lambda^{opv}$ Object Velocity & $0$ & $-$ & $0$ \\ \hline
$\lambda^{orv}$ Object Angular Velocity & $0$ & $-$ & $0$ \\ \hline
$\lambda^{rel}$ Relative Motion & $20$ & $-$ & $20$ \\ \hline
$\boldsymbol{\lambda^{cg}}[0]$ Ball-Hands Contact & $5$ & $-$ & $-$ \\ \hline
$\boldsymbol{\lambda^{cg}}[1]$ Ball-Body Contact & $5$ & $-$ & $-$ \\ \hline
$\boldsymbol{\lambda^{cg}}[2]$ Body-Hands Contact & $5$ & $-$ & $-$ \\ \hline
\end{tabular}%
}
\label{tab: sm reward weight}
\end{table}

\begin{table}[t]
\centering
\caption{Reward weights of DM in different tasks. DM related settings are tested only on Locomotion and BallPlay tasks.}
\resizebox{0.85\linewidth}{!}{%
\begin{tabular}{|l|c|c|}
\hline
\textbf{Parameter} & \textbf{BallPlay-M} & \textbf{Locomotion} \\ \hline
$\lambda^{p}$ Position & $40$ & $40$ \\ \hline
$\lambda^{r}$ Rotation & $2$ & $2$ \\ \hline
$\lambda^{rv}$ Rotation Velocity \quad\quad\quad\quad& $0.1$ & $0.1$ \\ \hline
$\lambda^{op}$ Object Position & $40$ & $-$\\ \hline
\end{tabular}%
}
\label{tab: dm reward weight}
\end{table}

\subsection{History Encoder Pre-training}
We present a self-supervised pre-training approach for the History Encoder (HE) that enables effective learning of temporal dependencies. The pre-training process follows a behavioral cloning paradigm, where we jointly train an encoder $\boldsymbol{\theta}$ to generate compact historical embeddings and an state predictor $\boldsymbol{\psi}$ to model state transitions. 
The encoder $\boldsymbol{\theta}$ consists of three convolutional layers followed by a fully connected layer, while the predictor $\boldsymbol{\psi}$ employs a three-layer MLP structure similar to the policy network.

Given a demonstration dataset, we randomly sample state trajectories $\{\hat{\boldsymbol{s}}_{t-k}, ..., \hat{\boldsymbol{s}}_{t+1}\}$ as reference sequences with their corresponding condition labels $\boldsymbol{c}$. The History Encoder $\boldsymbol{\theta}$ processes a sequence of $k$ historical states to generate a $\mu$-dimensional embedding $\boldsymbol{h}_t$:
\begin{equation}
\boldsymbol{h}_t = \boldsymbol{\theta}(\boldsymbol{s}_{t-k}, ..., \boldsymbol{s}_{t-1})
\end{equation}

This historical context is concatenated with the current state $\hat{\boldsymbol{s}}_t$ and condition $\boldsymbol{c}$, then passed to a state transition predictor $\boldsymbol{\psi}$, which estimates the next state:
\begin{equation}
\boldsymbol{s}_{t+1} = \boldsymbol{\psi}([\boldsymbol{c}, \hat{\boldsymbol{s}}_t,\boldsymbol{h}_t])
\end{equation}

The training objective combines state prediction accuracy with an embedding regularization term:
\begin{equation}
\mathcal{L} = \lambda_{a}|\hat{\boldsymbol{s}}_{t+1} - \boldsymbol{s}_{t+1}|^2 + \lambda_{b}|\boldsymbol{h}_t|^2
\end{equation}
where $\lambda_{a}=1$ and $\lambda_{b}=10^{-5}$ are hyperparameters controlling the balance between prediction accuracy and embedding regularization. Both $\boldsymbol{\theta}$ and $\boldsymbol{\psi}$ are optimized during pre-training through this objective. The predictor $\boldsymbol{\psi}$ effectively approximates the combined behavior of the policy and physical simulator, ensuring that successful convergence during pre-training indicates the HE has learned meaningful temporal representations.

\section{Hyperparameters}
\label{hyperparameters}
The hyperparameters for policy training are detailed in Tab.~\ref{tab: params_training}. Additionally, Tab.~\ref{tab: params_augmentation} presents the data augmentation hyperparameters, Tab.~\ref{tab: sm reward weight} and Tab.~\ref{tab: dm reward weight} displays the reward weights for SM and DM respectively.

\end{document}